\documentclass[a4paper,fleqn]{cas-dc}

\usepackage[numbers,sort&compress]{natbib}
\usepackage[linesnumbered,commentsnumbered,ruled,vlined]{algorithm2e}

\def\tsc#1{\csdef{#1}{\textsc{\lowercase{#1}}\xspace}}
\tsc{WGM}
\tsc{QE}
\tsc{EP}
\tsc{PMS}
\tsc{BEC}
\tsc{DE}

\begin{document}

\let\ref\Cref 		
\let\eqref\Cref 	
\let\autoref\Cref 	
\let\WriteBookmarks\relax
\def\floatpagepagefraction{1}
\def\textpagefraction{.001}
\shorttitle{}
\footmarks{
A version of this work was published in
Engineering Science and Technology, an International Journal, 61 (2025), 101920.\\
The final published version is available at
\url{https://doi.org/10.1016/j.jestch.2024.101920}.\\
This manuscript version is made available under the CC BY-NC-ND 4.0 license
(\url{https://creativecommons.org/licenses/by-nc-nd/4.0/}).
}

\bookmark[named = FirstPage]{Sample Selection Using Multi-Task Autoencoders in Federated Learning with Non-IID Data} 
\title [mode = title]{Sample Selection Using Multi-Task Autoencoders in Federated Learning with Non-IID Data}




\author{Emre ARDIÇ}[type=editor, 
                        auid=000,bioid=1,
                        ]
\ead{eardic@gtu.edu.tr}

\credit{Conceptualization, Methodology, Software, Validation, Formal analysis, Investigation, Resources, Data curation, Writing–original draft, Writing–review \& editing, Visualization, Project administration}

\address{Gebze Technical University, Department of Computer Engineering,
	Gebze, Kocaeli 41400, Turkey}

\author{Yakup GENÇ}[
   ]
\ead{yakup.genc@gtu.edu.tr}

\credit{Conceptualization, Methodology, Writing–original draft, Writing–review \& editing, Supervision}

\begin{abstract} 
Federated learning is a machine learning paradigm in which multiple devices collaboratively train a model under the supervision of a central server while ensuring data privacy. However, its performance is often hindered by redundant, malicious, or abnormal samples, leading to model degradation and inefficiency. To overcome these issues, we propose novel sample selection methods for image classification, employing a multi-task autoencoder to estimate sample contributions through loss and feature analysis. Our approach incorporates unsupervised outlier detection, using one-class support vector machine (OCSVM), isolation forest (IF), and adaptive loss threshold (AT) methods managed by a central server to filter noisy samples on clients. We also propose a multi-class deep support vector data description (SVDD) loss controlled by a central server to enhance feature-based sample selection. We validate our methods on CIFAR10 and MNIST datasets across varying numbers of clients, non-IID distributions, and noise levels up to 40\%. The results show significant accuracy improvements with loss-based sample selection, achieving gains of up to 7.02\% on CIFAR10 with OCSVM and 1.83\% on MNIST with AT. Additionally, our federated SVDD loss further improves feature-based sample selection, yielding accuracy gains of up to 0.99\% on CIFAR10 with OCSVM. These results show the effectiveness of our methods in improving model accuracy across various client counts and noise conditions.
\end{abstract}


\maketitle
\section{Introduction}

\begin{figure}[pos=t]
	\centering
	\includegraphics[width=0.90\columnwidth]{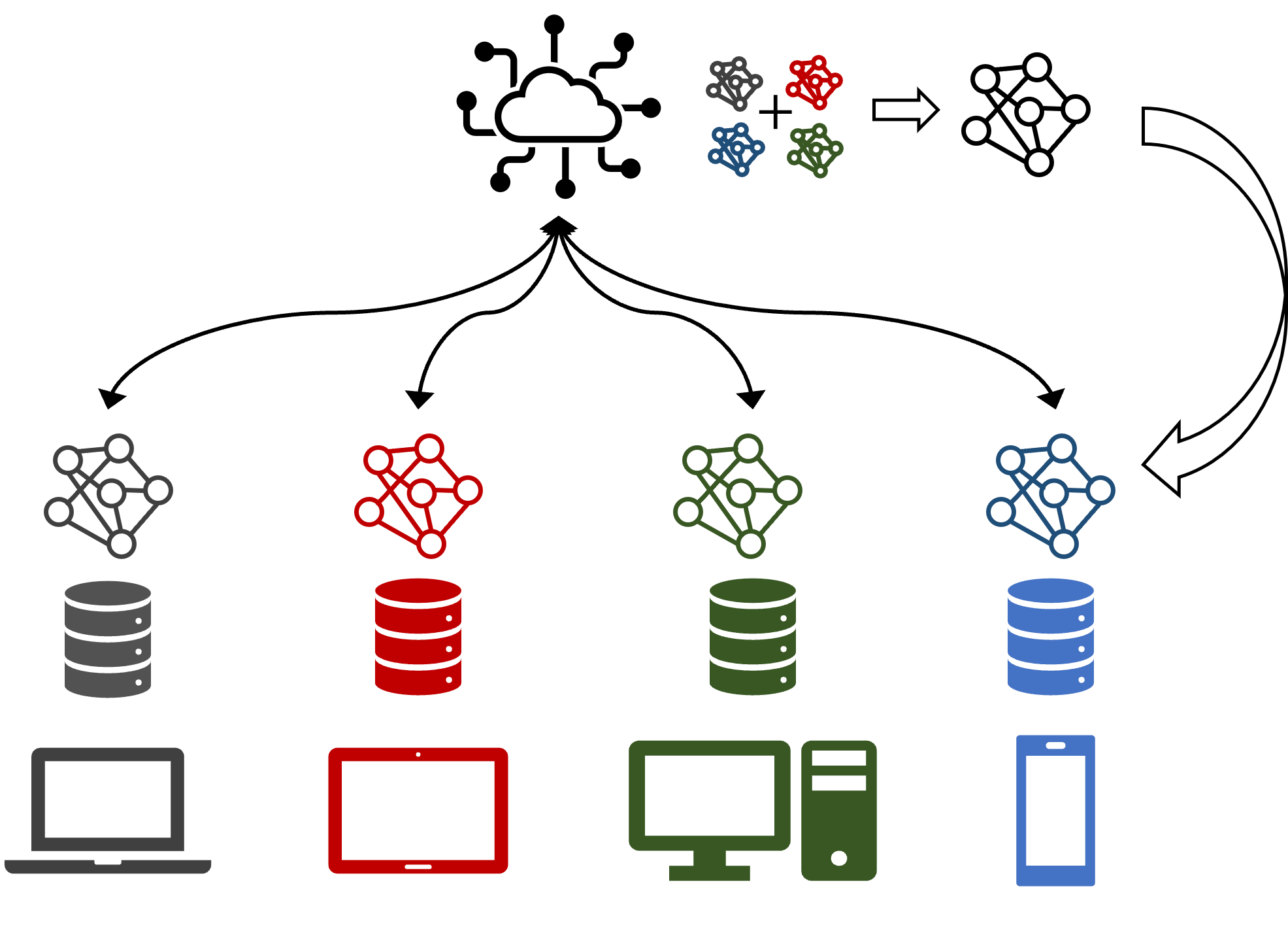}
	\caption{The typical training process of federated learning with various types of clients and a single server. Best viewed in color.}
	\label{fl_op}
\end{figure}

Modern distributed networks, including smartphones, wearable devices, and self-driving vehicles, produce a large amount of data. As the computing power, storage capacity, and battery life of these devices improve, local data storage and processing become more feasible and secure. This has led to a growing interest in Federated Learning (FL), a novel machine learning approach where many clients, such as individual mobile phones or tablets, collaboratively train a model under the management of a central server while keeping the training data decentralized \cite{mcmahan2017com}. The raw data generated by each client remains local, and only model updates are transferred to a global server to optimize a learning objective, thereby increasing data privacy as illustrated in \ref{fl_op}. Federated learning is well-suited for mobile devices, IoT systems, and institutions, particularly for tasks like face, fingerprint, and speech recognition, as well as predictive text in smartphones, where data privacy is crucial. Organizations like hospitals that keep sensitive private data, such as lung scans and brain MRIs, can be seen as clients in this context.

The key problems of FL are communication overhead, device heterogeneity, and statistical heterogeneity of data \cite{Li2020}. Variations in communication and computational capabilities of devices make it crucial to reduce the number of communication rounds and the size of transmitted data. Moreover, the statistical heterogeneity of data poses a significant challenge, as each device generates training data based on its unique local environment and usage patterns, leading to substantial differences in the size and distribution of datasets. This non-independent and identically distributed (non-IID) data generation process introduces bias into the training, leading to slow convergence or training failures \cite{sattler2019robust}. Effective sample selection can mitigate these challenges by filtering out abnormal samples on clients, resulting in more accurate model updates, faster convergence, and reduced communication costs \cite{li2021sample}.

Estimating sample contribution to model performance is a fundamental yet underexplored problem in federated learning \cite{li2021sample}. Potential use cases are anomaly detection \cite{novoa_anomalyae_2023fl}, robust learning \cite{Li2020AnomFL, iofl}, and client valuation \cite{li2021sample, Zhao2021}. By estimating the value of individual samples, it becomes possible to identify abnormal samples by closely examining low-quality data. Similarly, the value of each client can be calculated based on the quality of its samples, allowing for a more refined assessment of client contributions to the overall model. In this study, we focus on computing sample contributions locally on each client, allowing us to skip malicious, noisy, or abnormal samples during training \cite{li2021sample, eardic_fl}. This not only maintains data quality but also reduces the influence of outliers and protects against adversarial attacks, ultimately leading to more accurate and robust models. In this context, autoencoders are especially useful as they can effectively capture the underlying data distribution by reconstructing inputs. This allows for the identification of outliers through high reconstruction errors, making it easier to detect outliers, noisy or malicious samples \cite{kea2023enhancing_anomalyae_fl, nardi2022anomaly_uodfl, cheng2021improved}.

Our motivation arises from the inefficiencies of current data valuation methods in FL. Although Shapley Value (SV) provides theoretically robust estimations, its high computational cost makes it impractical for clients with limited resources \cite{wang2020principled}. Therefore, existing approaches often rely on loss or gradient norm to estimate sample contributions \cite{li2021sample, shin2022fedbalancer}. However, these methods have been relatively underexplored in large-scale, non-IID settings involving a high number of clients (up to 1000), particularly in unsupervised settings. Moreover, the use of autoencoders in FL, especially under non-IID conditions and at scale, has received relatively little attention \cite{nardi2022anomaly_uodfl}. To address these gaps, we propose a novel approach that combines classification and reconstruction losses. This dual-loss strategy improves the detection of abnormal samples by leveraging classification loss to identify mislabeled data and reconstruction loss to detect structural abnormalities, leading to more efficient and robust anomaly detection.

In this paper, we present novel sample selection methods specifically tailored for federated learning to improve the accuracy of models by filtering abnormal samples on clients. Our approach uses a Multi-Task Autoencoder (MTAE) architecture that serves two purposes: Image Classification (IC) and Reconstruction (IR). This design allows us to estimate the contribution of each sample by analyzing both IR and IC losses and outlier detection through feature analysis. Before local training on clients, we utilize One-Class Support Vector Machine (OCSVM) \cite{libsvm} and Isolation Forest (IF) \cite{liu2008isolation} models to identify and eliminate outlier samples. On the central server, these models are periodically trained using the features or losses gathered from the clients. In addition, we propose an adaptive threshold (AT) method, inspired by FedBalancer \cite{shin2022fedbalancer}, to filter outliers on clients using a global threshold based on a weighted sum of IR and IC losses, modifying the original by removing deadline control and changing the sampling strategy. We also propose a federated variant of the Support Vector Data Description (SVDD) loss \cite{cheng2021improved}, added as a regularization term to the main loss, to enhance feature-based sample selection on clients by manipulating the feature space. In experiments, we assess our sample selection methods on non-IID datasets across varying client counts and noise types, including closed-set and open-set noise. 

The summary of contributions is as follows:

\renewcommand\labelitemi{\small$\bullet$} 
\begin{itemize}
\itemsep=-1pt 		
\itemindent=-3pt 	
\item A multi-task autoencoder architecture is proposed for image classification and reconstruction tasks, showing its usefulness in improving accuracy through outlier detection on clients via loss and feature analysis.
\item A novel unsupervised outlier detection strategy, managed by a central server, is proposed for non-IID FL, along with a new variant of the FedBalancer, proving its effectiveness across different client counts and noise types.
\item A multi-class federated SVDD loss is proposed to improve feature-based outlier detection on clients, showing its effectiveness on two different datasets with open-set noise.
\end{itemize}

\section{Related Works}

The FL literature on data and client valuation using SV is limited \cite{wang2020principled, wang2019measure, Zhao2021}. For example, T. Wang et al. propose Federated Shapley Value \cite{wang2020principled} algorithm, a variant of SV, which is a popular method in cooperative game theory to compute total gains generated by a set of players. They consider FL to be a cooperative game between various clients or data sources. G. Wang et al. measure the contributions of participants in both vertical and horizontal FL \cite{wang2019measure}. They use SV for vertical FL and a deletion method for horizontal FL to calculate grouped feature and instance importance, respectively. Ardic et al. estimate sample contributions on clients by using leave-one-out and SV methods, which require retraining of local models multiple times \cite{eardic_fl}. However, these methods are computationally expensive and hard to use in FL.

Another way to estimate sample contribution is by using loss or gradient norm, calculated through a forward pass of the model. \cite{li2021sample, katharopoulos2018not, shin2022fedbalancer}. While the SV methods are accurate, loss-based methods are much faster, given the limited computing power of FL clients. Li et al. use gradient norm to estimate the importance of samples by using a single forward \cite{li2021sample}. They compute the importance of a client by summing the importance of its local samples. During training, the central server selects a group of important clients and their high-quality samples for each round to enhance accuracy. However, gradient norm calculation is more costly in terms of time and space compared to loss calculation. Shyn et al. propose a simple accuracy approximation model to estimate client contributions by using only data size \cite{Shyn2021}. Tuor et al. propose a benchmark model trained on a small benchmark dataset to overcome noisy and irrelevant data in FL \cite{tuor2021overcoming}. Consequently, relying on sample count is unreliable, as it overlooks data quality, and the benchmark models are task-specific and hard to generalize.

Anomaly detection is an emerging research area in FL, drawing considerable attention due to its crucial applications in various domains such as cybersecurity, healthcare, and industrial systems \cite{trafficanom_yiluan2024, tran2022improved_sensoranom, kea_poweranom}. A prominent method involves using autoencoders, which makes it easier to identify noisy and malicious samples through IR loss \cite{Li2020AnomFL, li2019abnormalcli, novoa_anomalyae_2023fl, kea2023enhancing_anomalyae_fl, bhat2023anomaly}. Nardi et al. propose an anomaly detection method for unsupervised tasks in decentralized settings. Their approach first groups clients into communities with similar data patterns and trains a shared autoencoder-based anomaly detection model for each community. \cite{nardi2022anomaly_uodfl}. However, their method is not suitable for a large number of clients and has only been tested on fully connected networks. Bhat et al. propose a performance-based parameter aggregation method in FL, assigning weights to clients' models based on F1 scores to improve anomaly detection and reduce the impact of low-performing clients \cite{bhat2023anomaly}. However, their method does not consider non-IID data distributions, has not been tested with a large number of clients, and requires a test dataset on each client to compute F1 scores.

\section{Methods} \label{sec_methods}

In this section, we start by clearly defining federated learning and providing a detailed step-by-step explanation of the training process. Next, we introduce commonly used datasets for image classification and delve into various types of noise that may be encountered during training. Then, we introduce our MTAE architecture, which is tailored for both image classification and reconstruction tasks. Finally, we present our sample selection strategies, specifically designed for federated learning, based on loss and features.

\subsection{Federated Learning}

We consider a typical synchronous FL scenario with two types of elements: a server and $N$ clients, denoted as $\mathcal{N}=\{\zeta_1, \dots, \zeta_N\}$. Each client $\zeta_i$ has a local dataset $D_i=\{(x_{ij},y_{ij})\}_{j=1}^{m_i}$, where $x_{ij}$ represents $j$-th local sample on client $i$, $y_{ij}\in[c_1,\dots,c_k]$ is the corresponding label, and $m_i = |D_i|$ is the total number of local samples. The main objective is to train a single global model with the data generated by a large number of clients. In this process, only model updates or gradient information is periodically transferred between the server and clients. We employ the FedAvg algorithm to aggregate local models that are collected from remote devices in each round \cite{mcmahan2017com}. As shown in \ref{fl_op}, a central server manages the training process over rounds by repeating the following steps:

\begin{enumerate}[(1)]
\item \textit{Client selection:} The central server selects clients from a set of active clients by checking specific conditions such as battery level, connection quality, and CPU usage in order to avoid impacting the usage or functionality of the device.
\item \textit{Broadcast:} The selected clients download the current model weights and training configuration from the central server.
\item \textit{Local update:} Each selected client computes an update to the model by using its local data.
\item \textit{Aggregation:} The server collects an aggregate of the client updates. In this state, various methods like lossy compression and differential privacy can be integrated to reduce communication costs and preserve privacy.
\item \textit{Global update:} The server locally updates the global model based on the aggregated update computed from the selected clients in the current round.
\end{enumerate}

Let $\ell(f(x_{ij}; w), y_{ij})$ be the loss function evaluated on a sample $(x_{ij}, y_{ij}) \in D_i$, where $x_{ij}$ is an input sample, $y_{ij}$ is the ground-truth label, and $w$ denotes the model parameters. The model is defined by $f(x_{ij}; w)$, indicating that the model is a function of the input $x_{ij}$ parameterized by $w$. The overall loss $\mathcal{L}_i(w)$ at a client $\zeta_i$ is defined in Eq. \ref{eq_fiw}. During the training, each client tries to minimize its local loss function $\mathcal{L}_i(w)$, and depending on the application, modifications can be made to the mathematical model.

\begin{equation}
\mathcal{L}_i(w) = \frac{1}{m_i}\sum_{j=1}^{m_i}\ell(f(x_{ij}; w), y_{ij})
\label{eq_fiw}
\end{equation}
\smallskip

The primary goal during the federated learning process is to minimize the objective function as defined in Eq. \ref{eq_fl} \cite{Li2020}. This objective function is solved in a distributed fashion where each client performs multiple epochs of training using stochastic gradient descent (SGD) on its local model.

\begin{equation}
\hat{w}=\arg\min_w\frac{\sum_{i=1}^{N}m_i\mathcal{L}_i(w)}{\sum_{i=1}^{N}m_i}
\label{eq_fl}
\end{equation}

\subsection{Datasets \& Noise Types} \label{sec_methods_dataset}

\begin{figure}[pos=t]
	\centering
	\includegraphics[width=1\columnwidth]{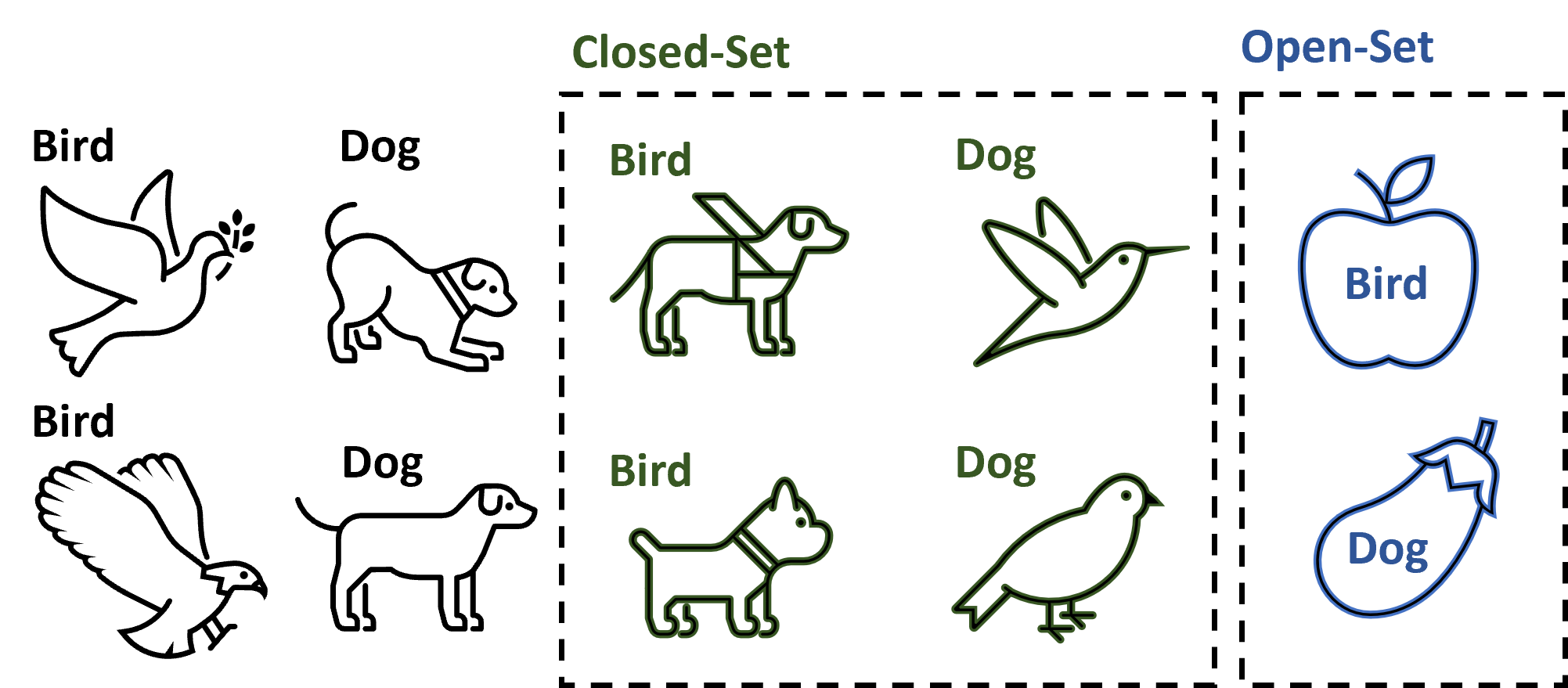}
	\caption{The illustration of open-set and closed-set label noise. Consider a noisy training dataset of bird and dog images. The images of dogs that are labeled as birds or vice-versa belong to the closed-set noise. Images that are neither birds nor dogs belong to the open-set noise. Best viewed in color.}
	\label{fig_noise}
\end{figure}

 In federated learning, clients may intentionally or unintentionally generate noisy, redundant, or abnormal samples that can disrupt the training process. In this work, we consider that the clients are trusted and cooperative. In a typical machine learning training, we have all the data at our disposal. This allows us to analyze and clean up the dataset before the training to improve model accuracy. In federated learning, however, we do not have permission to access clients' datasets. Thus, it is a challenging task to select only valuable samples during the training. To simulate this scenario, we inject open-set and closed-set noise into the training datasets and distribute them to clients in a non-IID fashion. These noise types illustrated in \ref{fig_noise} are explained as follows:

\begin{enumerate}[(1)]
\item \textit{Closed-set noise:} A data sample from one known class is incorrectly labeled as another known class.
\item \textit{Open-set noise:} A data sample from an unknown class is incorrectly labeled as a known class.
\end{enumerate}

In this study, we introduce closed-set noise by swapping labels within the training dataset and open-set noise by substituting images with random images from other datasets. We train and test our FL models by using MNIST \cite{deng2012mnist} and CIFAR10 \cite{cifar10} datasets. MNIST, widely used in machine learning, consists of 28x28 grayscale images of handwritten digits collected from 1,000 users. CIFAR10, another prominent dataset, contains 32x32 color images of 10 different objects, such as cars, birds, and trucks. For open-set noise, we employ datasets such as Street View House Number (SVHN) \cite{svhn}, ImageNet32 \cite{imagenet32}, and Extended MNIST (EMNIST) \cite{cohen2017emnist}. We use the federated version of the EMNIST dataset as suggested by Google AI \cite{reddi2020adaptive}. SVHN is obtained from house numbers in Google Street View images and contains 32x32 cropped digits of 10 classes. EMNIST extends MNIST by including upper-case and lower-case English characters. Additionally, ImageNet32, a downscaled version of ImageNet, consists of 32x32 color images spanning 1,000 classes. More detailed descriptions of these datasets are provided in \ref{table:datasets}.

\begin{table}[width=1\linewidth,cols=5,pos=b]
	\caption{The details of datasets used in this study, including training datasets CIFAR10 and MNIST, and others used for open-set noise.}
    \label{table:datasets}
	\begin{tabular*}{\tblwidth}{@{} LLLLL@{} }
		\toprule
        Dataset & Image Size & Train Examples & Test Examples & Classes \\
		\midrule
            CIFAR10     & 32x32 RGB  &	50,000    &	10,000  &	10    \\
            MNIST       & 28x28 Gray &	60,000    &	10,000  &	10    \\
            SVHN        & 32x32 RGB  &	73,257    &	26,032  &	10    \\
            EMNIST      & 28x28 Gray &	671,585   &	77,483  &	62    \\
            ImageNet32  & 32x32 RGB  &	1,281,167 &	50,000  &	1,000  \\
		\bottomrule
	\end{tabular*}
\end{table}

\subsection{Multi-Task Autoencoders} \label{sec_mtae}

Let $(x_j, y_j) \in D$ be a sample from dataset $D$, where $x_j$ is the input image, and $y_j$ is the corresponding label. A multi-task autoencoder (MTAE) comprising an encoder, a decoder, and a classifier can be defined as $f: x_j \rightarrow z_j$, $g: z_j \rightarrow \hat{x}_j$, and $h: z_j \rightarrow \hat{y}_j$. Here, the reconstructed input is $\hat{x}_j=g(f(x_j))$, the predicted label is $\hat{y}_j=h(f(x_j))$, and $z_j$ is the embedded representation of $x_j$ in the feature space $\mathcal{F}$. In this study, the reconstruction loss $\mathcal{L}_{\text{rec}}(x_j, \hat{x}_j)$ is measured using Mean Squared Error (MSE), and the classification loss $\mathcal{L}_{\text{cls}}(y_j,\hat{y}_j)$ is computed using Cross-Entropy (CE). The overall loss $\mathcal{L}$ in Eq.\ref{mtae_loss} is minimized ($\min_{f, g, h} \mathcal{L}$), jointly optimizing the reconstruction loss \(\mathcal{L}_{\text{rec}}\) and the classification loss \(\mathcal{L}_{\text{cls}}\). The hyperparameters $\lambda_{\text{rec}}$ and $\lambda_{\text{cls}}$ are crucial, as they control the influence of pixel noise and label noise on the model, respectively.

\begin{equation}
\mathcal{L}=\lambda_{\text{rec}} \mathcal{L}_{\text{rec}}(x_j, \hat{x}_j) + \lambda_{\text{cls}} \mathcal{L}_{\text{cls}}(y_j, \hat{y}_j)
\label{mtae_loss}
\end{equation}
\smallskip

\begin{figure}[pos=t]
	\begin{subfigure}{.5\textwidth}
		\centering
        \includegraphics[width=1\columnwidth]{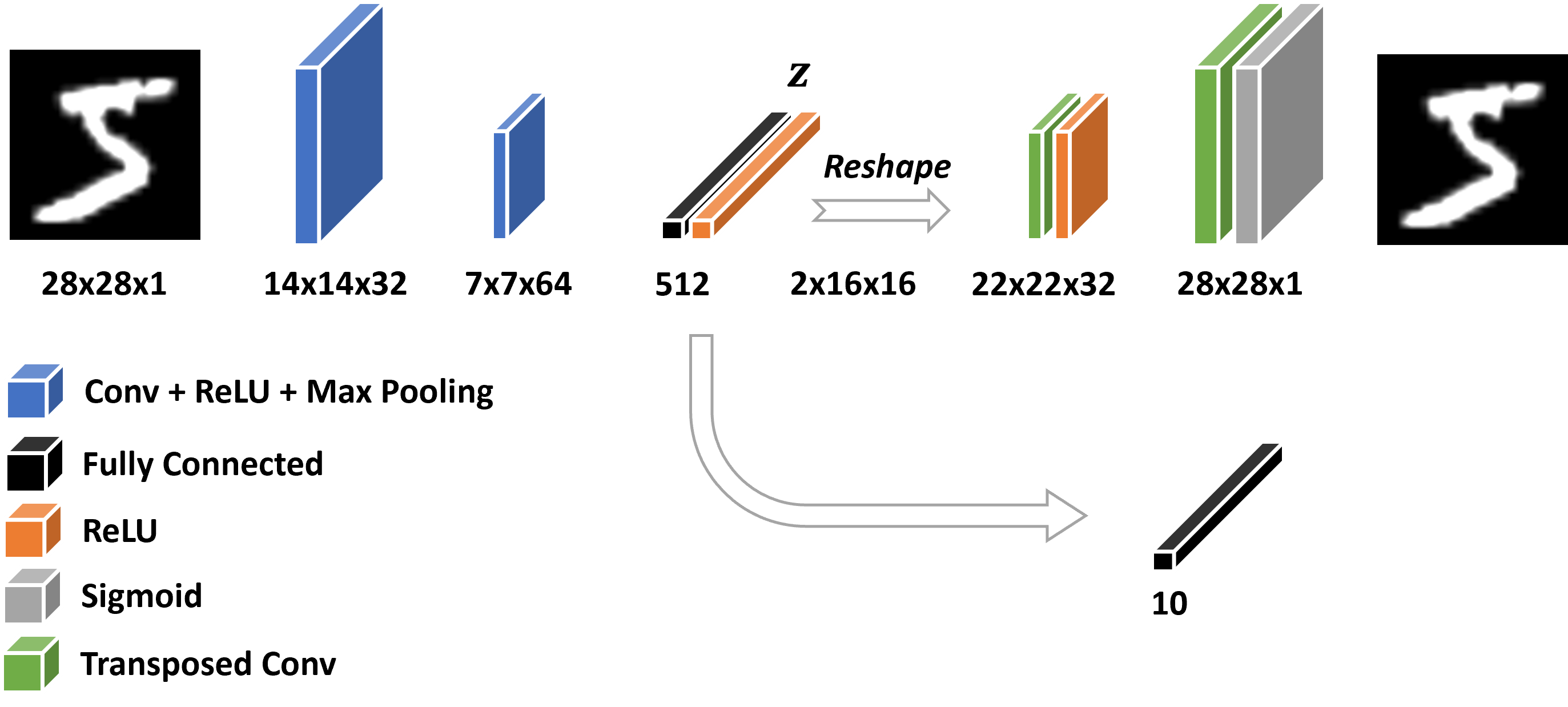} 
		\caption{}  				
		\label{model_mnist}
	\end{subfigure}
	\begin{subfigure}{.5\textwidth}
		\centering
        \includegraphics[width=1\columnwidth]{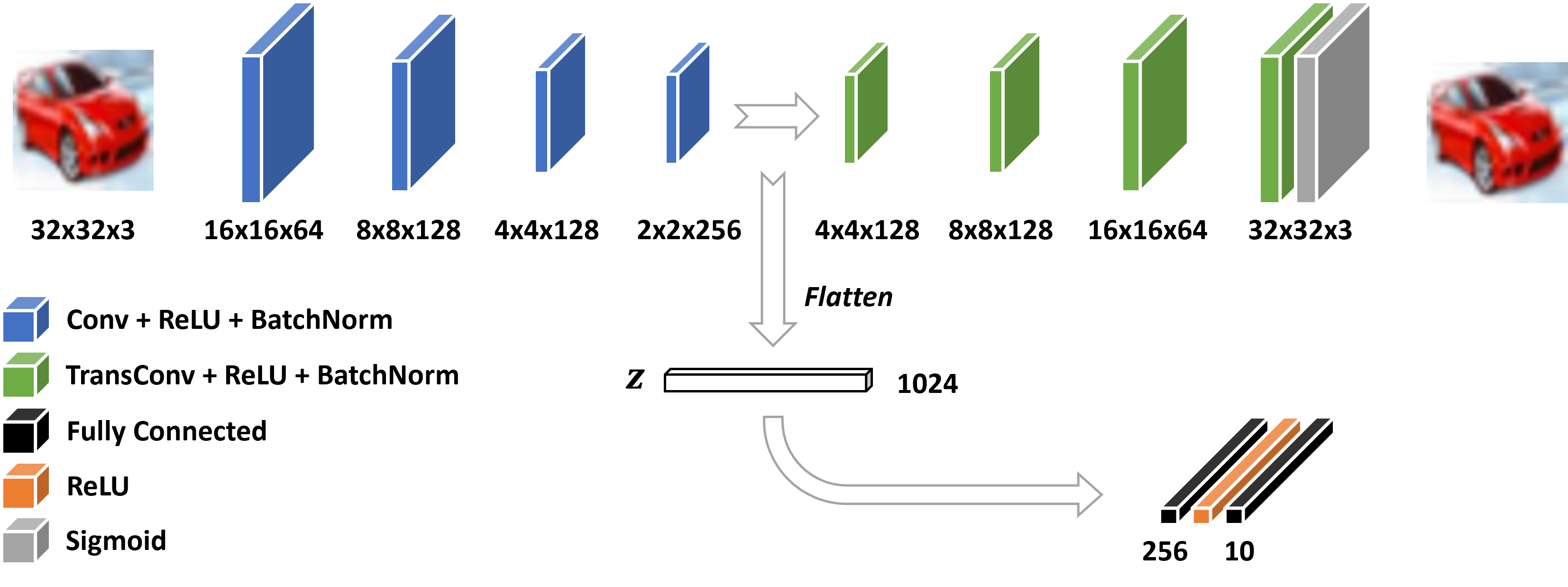}
		\caption{} 					
		\label{model_cifar}
	\end{subfigure}
	\caption{The MTAE architectures designed for MNIST (a) and CIFAR10 (b) datasets. Best viewed in color.}
	\label{model_all}
\end{figure}

The MTAE model designed for the MNIST is shown in \ref{model_mnist}. We use an encoder based on a two-layer CNN with 32 and 64 filters, similar to the architecture employed in FedAvg \cite{mcmahan2017com}. The kernel size, stride, and padding parameters of the CNN layers are set to 5, 1, and 2. Following each convolution layer, the encoder applies the Rectified Linear Unit (ReLU) activation function and a max-pooling layer with a kernel size of 2x2 and a stride of 2. The embedding $z_j$ size of the model is 512. The classification layer consists of a single Fully Connected (FC) layer responsible for learning the 10 classes from this embedding $z_j$. We use a decoder based on a two-layer transposed CNN with 32 and 1 filters, each having a kernel size of 7x7 and a stride of 1. The first layer uses a ReLU, while the final layer employs a sigmoid activation function, yielding output values within the range of 0 to 1.

The MTAE model designed for the CIFAR10 dataset with 32x32 color images is presented in \ref{model_cifar}. We use an encoder with a four-layered CNN network consisting of 64, 128, 128, and 256 filters, each having a kernel size of 3x3, a stride of 2, and a padding of 1. After each convolutional layer, we apply ReLU activation function and batch normalization. We use two FC layers in the classification layer and apply a ReLU activation function to the first one. The last FC layer is responsible for learning 10 classes. The embedding $z_j$ size of the model is 1024, and it serves as input to the classification layer. The decoder has four-layered transposed CNN with 128, 128, 64, and 3 filters, each having a kernel size of 3x3, a stride of 2, a padding of 1, and an output padding of 1. Following each layer, we apply ReLU activation function and batch normalization. We use the sigmoid activation function in the final layer of the decoder, yielding output values between 0 and 1.

\subsection{Sample Contribution} \label{sec_methods_ss}

The contribution of a sample $x_j \in D$ to a global model $A$ can be defined as a function of $D$ and $A$, denoted as $\phi_j(D, A)$ or simply $\phi_j$. In this paper, the aim is to identify and remove malicious, abnormal, or redundant samples by estimating $\phi_j$ values of the local samples efficiently. To achieve this, we propose loss-based approaches where the loss values, considered as $\phi_j$, are computed by a single forward process in clients. A high loss value indicates that the corresponding sample is either valuable or an outlier. Conversely, a sample with a low loss is redundant or has already been effectively learned by the model. The models trained solely with cross-entropy loss are sensitive to both label and pixel noise. Similarly, relying on image reconstruction error as $\phi_j$ is not reliable due to the ability of autoencoders to generalize exceptionally well on the training data, to the point of accurately reconstructing both normal and anomalous data. The loss values are typically higher for malicious, abnormal, or noisy images compared to clean ones because the model tends to generate higher losses for outlier samples \cite{cheng2021improved}. However, this assumption does not always hold true, particularly in unsupervised anomaly detection, where the training data itself may include anomalies. To better measure the $\phi_j$ value of each sample during training, we propose an MTAE architecture with image classification and reconstruction. This design, as shown in \ref{model_all}, helps us to identify label noise through CE loss and pixel noise through MSE loss. This approach also allows us to use autoencoder features for $\phi_j$ analysis.

\subsection{Adaptive Thresholding} \label{sec_at}

We propose an adaptive threshold method for loss-based sample selection in FL, modifying the sample selection module of FedBalancer by removing its deadline control \cite{shin2022fedbalancer}. Our approach uses a loss function that combines image reconstruction and classification losses into a weighted sum, with weights defined in \ref{sec_train_conf}. The sample selection procedure (\textit{SelectSample}) is defined in Algorithm 1, where $D_i$ is the local dataset of client $i$ and $loss_i$ is the losses of local samples. In the $R$-th round, client $i$ selects its samples based on their loss values, dividing them into Under-Threshold ($\text{UT}^i_R$) and Over-Threshold ($\text{OT}^i_R$) sets using the threshold $lt_R$. The samples in $\text{OT}^i_R$ are considered outliers or important samples, which are selected randomly (Line 24), with the selection amount determined by $p \in [0.0, 1.0]$. In contrast, the samples in $\text{UT}^i_R$ are presumed to be normal and included in the training.

At the beginning of training, the loss values of all samples are nearly identical, so the loss threshold $lt_R$ is gradually increased to allow clients sufficient time to learn. The threshold selection module (\textit{CalculateLt}) in Algorithm 1 computes an optimal threshold on the server to effectively distinguish between clean and noisy samples. As training progresses and the loss distribution of samples changes, the module adapts to the new distribution. The server collects metadata containing loss statistics from the clients at the end of each round. Specifically, client $i$ in round $R$ provides the lowest ($LLow_R^i$) and highest ($LHigh_R^i$) loss values for its samples, which are aggregated into server-side lists $LLow_R$ and $LHigh_R$. The module then computes a new threshold ($lt_{R+1}$) by estimating the range of losses across clients ($ll$ and $lh$) and performing a linear interpolation (Line 5): $lt_{R+1} = ll + (lh - ll) \cdot ltr$. 

The loss threshold ratio ($ltr$) starts at 0 and is gradually increased by the loss step size ($lss$) to remove noisy samples as training progresses. As defined in the \textit{ControlLtr} function of Algorithm 1, the benefit of the current threshold is evaluated in each round according to statistical utility: $U_R = (LossSum_R) / L_R$, where $LossSum_R$ is the total loss of the selected samples and $L_R$ is the total number of selected samples. If the recent rounds have a lower average loss than the past ones, training is assumed stable, and $ltr$ is increased (Line 12); otherwise, it is decreased (Line 14). This control occurs every $t_w$ rounds (Line 10) to find the optimal threshold for separating normal and noisy samples. If the threshold is too high, the noisy samples increase, and $ltr$ is reduced.

\subsection{Unsupervised Outlier Detection} \label{sec_uod}

\begin{algorithm}[t]
\small
\SetKwProg{CalculateLt}{CalculateLt}{($LLow_R$, $LHigh_R$, $ltr$)}{}
\SetKwProg{ControlLtr}{ControlLtr}{($U$, $LossSum_R$, $L_R$, $ltr$, $lss$, $t_w$, $R$)}{}
\SetKwProg{SelectSample}{SelectSample}{($D_i$, $loss_i$, $lt_R$, $p$)}{}
\caption{The adaptive threshold algorithm}
\label{alg:adaptivelt}
\textbf{Server:}\\
\CalculateLt{}{
    $ll = min(LLow_R)$\\
    $lh = mean(LHigh_R)$\\
    $lt_{R+1} = ll + (lh - ll) \cdot ltr$\\
    \Return $lt_{R+1}$
}
\ControlLtr{}{
    $U_R = LossSum_R / L_R$\\
    U.add($U_R$)\\
    \If{$R \bmod t_w \equiv 0$}{
        \eIf{$\sum U(R-2t_w:R-t_w) > \sum U(R-t_w:R)$}{
            $ltr = min(ltr + lss, 1)$\\
        }{
            $ltr = max(ltr - lss, 0)$\\
        }
    }
    \Return $ltr$
}
\textbf{Client:}\\
\SelectSample{}{
    $OT^i$, $UT^i = []$\\
    \For{$x = 0$ \KwTo $|D_i|-1$}{
        \If{$loss_i[x] \geq lt_R$}{
            $OT^i.\text{insert}(D_i[x])$\\
        }
        \Else{
            $UT^i.\text{insert}(D_i[x])$\\
        }
    }
    \Return $UT^i \cup \text{randSample}(OT^i, p \cdot |OT^i|)$
}
\end{algorithm}

In this study, we use unsupervised outlier detection methods, OCSVM \cite{libsvm} and IF \cite{liu2008isolation}, to identify outlier samples either within a 2D loss space created by weighted CE and MSE losses or within the feature space. Our sample selection algorithm using OCSVM and IF is defined in Algorithm 2. Here, $T$ denotes the total number of training rounds, $P$ is the number of clients per round, $\mathcal{N}$ represents the set of all clients, $\mathcal{P}_t$ represents the set of clients selected at a round, $E$ is the number of local training epochs, $B$ is the batch size, $\eta$ is the learning rate, $\mathcal{M}$ is the outlier detection model, $t_s$ specifies the round after which sample selection begins, and $t_w$ is the interval at which the outlier detection model is updated.

Instead of training the outlier detection model on each client, which adds extra computational cost, we train the model on the central server by using the losses or features collected from each client. Then, we broadcast the new model to clients along with the global FL model. Before the local training on clients, each client computes loss values (Line 18) and removes outlier samples (Line 20) from the local training by using the outlier detection model. The outlier detection model can be retrained every $t_w$ rounds (Line 13) rather than after each round to improve computational efficiency. The same process is also applied for feature-based sample selection. The difference is that we extract the embeddings of samples (Line 18) instead of computing losses, and the outlier detection is done in the feature space instead of the 2D loss space.

The sample selection process is started after a predefined warm-up round ($t_s$) during training to facilitate the learning of normal samples rather than memorizing abnormal ones. The choice of warm-up rounds is critical as prolonged training rounds make it hard to identify abnormal samples in the CE-MSE loss space due to possible overfitting. Moreover, in the feature-based sample selection, it is important to give the model enough time to learn an effective embedding space for distinguishing dissimilar samples.

\begin{algorithm}[t]
\small
\caption{The FedAvg with outlier detection}
\label{alg:uod}
\SetKwProg{FS}{Server}{($T$, $P$, $\mathcal{N}$, $E$, $B$, $\eta$, $t_s$, $t_w$):}{}
\FS{}{
    initialize global model parameters $w_0$ \\
    initialize outlier detection model $\mathcal{M}$ \\
    \For{$t = 0$ \KwTo $T - 1$}{
        select $P$ clients randomly as $\mathcal{P}_t \subset \mathcal{N}$ \\
        $m = 0$, $ssTrainSet = []$ \\
        \For{each client $i \in \mathcal{P}_t$}{
            $w^{i}_{t+1}, m_i, \kappa_i = \text{Client}(i, w_t, \eta, E, B, \mathcal{M}, t_s)$ \\   
            $ssTrainSet = ssTrainSet \cup \kappa_i$ \\
            $m \mathrel{+}= m_i$ \\
        }
        \If{$t \geq t_s \textbf{ and } (t \bmod t_w = 0$ \textbf{or} $t = t_s$)}{
            $\mathcal{M}.\text{fit}(ssTrainSet)$ \\
        }
        $w_{t+1} = \sum_{i \in \mathcal{P}_t} \cfrac{m_i w^{i}_{t+1}}{m} $ \\
    }
    \Return $w_T$
}
\SetKwProg{FC}{Client}{($i$, $w$, $\eta$, $E$, $B$, $\mathcal{M}$, $t_s$):}{}
\FC{}{
    $\kappa_i = []$ \\
    \If{$t \geq t_s$}{
        $\kappa_i = \text{ExtractLossesOrFeatures}(D_i, w)$ \\
    } 
    \If{$t > t_s$}{
        $D_i = \text{RemoveOutlierSamples}(D_i, \mathcal{M}(\kappa_i))$ \\
    }
    $m_i = |D_i|$ \\
    $\mathcal{B}_i$ = split $D_i$ into batches of size $B$ \\
    \For{$e = 0$ \KwTo $E - 1$}{
        \For{batch $(x, y) \in \mathcal{B}_i$}{
            $w \gets w - \eta \nabla \ell(f(x; w), y)$\\
        }
    }
    \Return $w, m_i, \kappa_i$
}
\end{algorithm}

\subsection{Federated Multi-Class SVDD Loss} \label{sec_fedsvdd}

An autoencoder with encoder and decoder can be defined as $f: x_j \rightarrow z_j$ and $g: z_j \rightarrow \hat{x}_j$, where $z_j$ is the embedded point of $x_j$ in the feature space $\mathcal{F}$, and $\hat{x}_j$ is the reconstructed input $x_j$. To improve outlier detection in the feature space $\mathcal{F}$, we propose multi-class deep SVDD loss as defined in Eq.\ref{svdd}. The focus is to find a good $f$ that makes embedded points $\{z_j\}_{j=1}^n$ more suitable for outlier detection in multi-class scenarios. The Deep SVDD loss $\mathcal{L}_{\text{reg}}$ modifies the feature space and is added as a regularization term to the autoencoder loss $\mathcal{L}$ as defined in Eq.\ref{mtae_loss}. The combined loss is $\mathcal{L}' = \mathcal{L} + \lambda_{\text{reg}} \mathcal{L}_{\text{reg}}$, where $\lambda_{\text{reg}}$ is a coefficient that controls the degree of feature space distortion. The objective is to learn a minimum hypersphere characterized by radius $R_i > 0$ and centroid $\mu_i \in \mathcal{F}$ for a specific class $i \in [1,k]$, where $k$ is the number of classes, $n$ is the total samples, $n_i$ is the number of samples in class $i$, and $\mathbb{1}_{y_j=i}$ is 1 if $y_j = i$ and 0 otherwise. This method aims to map the majority of normal data points inside the hypersphere while pushing anomalous data points outside the hypersphere.

\begin{equation}
\mathcal{L}_{\text{reg}} = \frac{1}{k} \sum_{i=1}^{k} \Big( R^2_i + \frac{1}{n_i} \sum_{j=1}^{n} \mathbb{1}_{y_j=i} \max\{0, \left\| f(x_j) - \mu_i \right\|^2 - R^2_i \} \Big)
\label{svdd}
\end{equation}
\smallskip

\begin{figure}[pos=t]
	\centering
	\includegraphics[width=1\columnwidth]{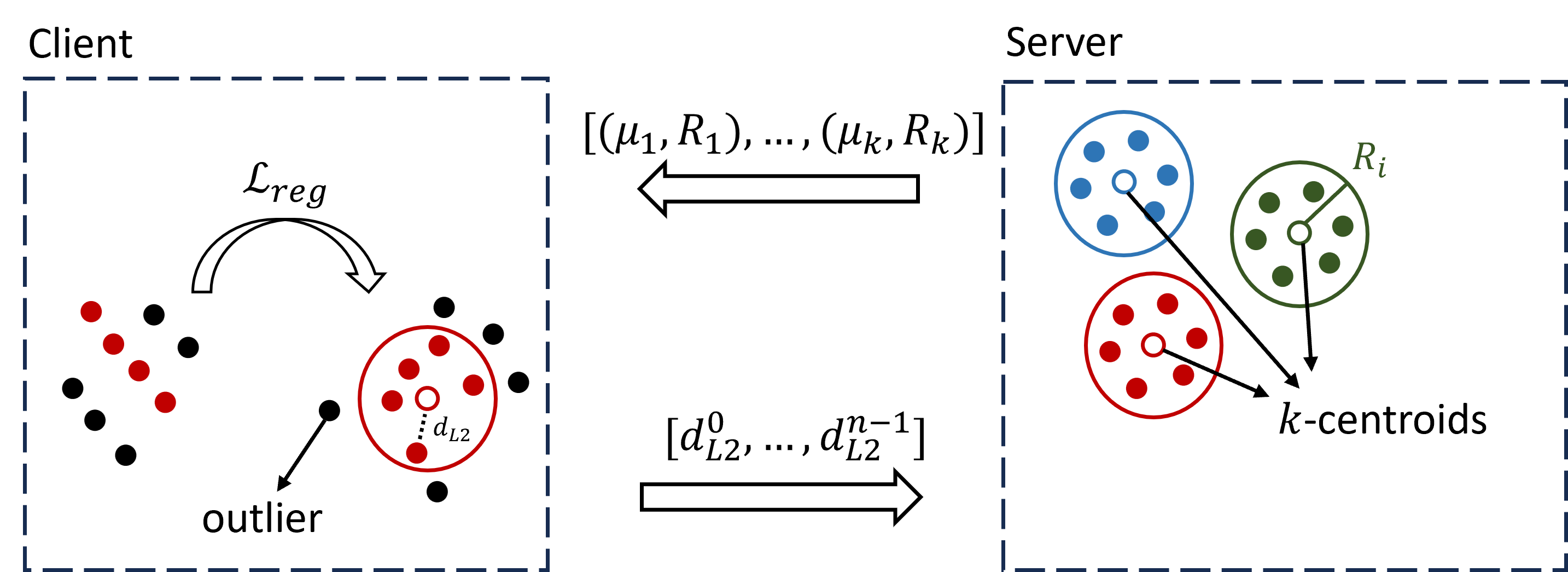}
	\caption{An overview of multi-class federated SVDD loss for a client and server where $n$ is the number of local samples and $d_{L2}$ represents L2 distance in the feature space. Best viewed in color.}
	\label{fed_svdd}
\end{figure}

Another contribution of this work is the adaptation of the Deep SVDD Loss \cite{cheng2021improved} to FL, as outlined in the computation steps presented in \ref{fed_svdd}. The server computes the class centroids $[\mu_1,\dots, \mu_k]$ and radii $[R_1,\dots, R_k]$ using a public test dataset once the global model achieves a predefined target, such as a specific validation accuracy, loss value, or training round. Subsequently, all centroids and radii are distributed to the selected clients during a round. On the clients, the proposed loss $\mathcal{L}_{\text{reg}}$ is computed and added as a regularization term to the original loss $\mathcal{L}$ during training. After local training, the clients calculate the Euclidean distances $[d_{L2}^j]_{j=1}^{n}$ between the embeddings of their samples and the class centroids $\mu_{1 \dots k}$, and then transmit these distances to the server. The server updates the radii $R_{1 \dots k}$ by calculating the $q$-th quantile of the distances, where $q$ is controlled by a hyperparameter $q=1-\nu$ that controls the fraction of data points considered as outliers. 

\section{Experiments and results} \label{sec_exp}

In this section, we present the comprehensive evaluation of our sample selection methods across varying numbers of clients and noise types on non-IID training datasets. Our primary objective is to show the efficacy of our sample selection methods in enhancing the global model's performance, leading to higher accuracy rates and faster convergence. Through a series of meticulously designed experiments, we explore diverse datasets injected with various types of noise and varying numbers of clients, comparing outcomes with and without our sample selection methods. Throughout the training process, we monitor the number of training rounds that yield the highest test accuracy, computed on the server using the global model, referred to as the "Best Round" in experiments. To fairly quantify the accuracy improvements achieved by our sample selection methods, we prefer the "Best Round" approach over the "Last Round" as the model accuracy fluctuates between rounds due to non-IID distribution and noise.

\subsection{Deployment} \label{sec_exp_deploy}

To assess our sample selection methods in an FL environment, we employ the FedML library, which offers simulations on a single machine, distributed computation, and edge device training, along with a generic programming interface featuring baseline implementations for optimizers, models, and datasets \cite{fedml2020}. By using this library, we develop a federated learning system with various numbers of clients on a single machine with NVIDIA RTX 3090 GPU, Ryzen 5900X CPU, 32GB RAM, and 1TB SSD by using Python 3.6, Scikit-Learn 0.24.2, PyTorch 1.8.2, and CUDA 11.1. We extend this library by implementing our sample selection methods, MTAE models, and both open-set and closed-set noise generators. We customize the single-process simulation module of FedML and use scikit-learn implementations of OCSVM and IF methods \cite{scikit-learn}.

\subsection{Datasets}

To illustrate the effectiveness of our sample selection methods, we introduce 40\% open-set or closed-set noise into the training datasets before the non-IID partitioning among clients. This high noise rate is chosen to demonstrate the benefits of the proposed sample selection methods more effectively, but it also makes it more challenging to achieve high accuracy. In our experiments, we employ the CIFAR10 and MNIST datasets for training while using ImageNet32, EMNIST, or SVHN datasets as sources of open-set noise. The open-set noise is applied in such a way that the number of images in each class is the same. In the CIFAR10 dataset experiments, we incorporate the ImageNet32 and SVHN datasets as open-set noise sources. To ensure a fair comparison, we exclude the common classes between CIFAR10 and ImageNet32 from the ImageNet32 dataset. Similarly, in the MNIST dataset experiments, the ImageNet32 and EMNIST datasets serve as open-set noise. Since the ImageNet32 dataset contains RGB images, we convert them to grayscale and resize them to 28x28 to match the MNIST dataset. Besides, we exclude the digit images in EMNIST for consistency. Given that the MNIST dataset contains handwritten digits from 1000 users, each user is treated as a client in our FL network, resulting in a naturally non-IID partitioning of the dataset. For experiments with fewer than 1000 clients, images from multiple users are randomly merged to achieve the desired client count. To apply closed-set noise to the training datasets, we randomly select training samples and assign incorrect labels to them. Further details about the datasets can be found in \ref{sec_methods_dataset}.

\subsection{Training Configuration} \label{sec_train_conf}

The MTAE models presented in \ref{sec_mtae} are trained for 1000 rounds, with the total number of clients varying. We use the FedAvg algorithm to aggregate local models in each round. The objective function, defined in Eq.\ref{mtae_loss}, is computed by using MSE and CE, where the loss weights are $\lambda_{\text{rec}}=1$ and $\lambda_{\text{cls}}=0.05$. The weight of CE is intentionally kept lower than that of MSE; otherwise, generated images get blurry, and the models cannot learn fine-grained details. For every 10 rounds of training, we compute train and test accuracies on the classification results as well as the Peak Signal-to-Noise Ratio (PSNR) and Structural Similarity Index (SSIM) metrics on the reconstructed images. Throughout training, we monitor the number of training rounds that yield the highest test accuracy. We employ a Stochastic Gradient Descent (SGD) optimizer with a learning rate of 0.1 and a weight decay of 0.001. The batch size is set to 64, and we perform 5 epochs of local training on clients. Our experiments are conducted with varying client numbers, including 1000, 200, 100, and 50, with corresponding client selections of 100, 20, 10, and 5 per round. We maintain a constant ratio of 0.1 between the number of clients per round and the total number of clients. The same random seed is used in all training sessions to ensure fair evaluation.

\subsection{Sample Selection} \label{sec_exp_ss}

The sample selection methods explained in \ref{sec_methods_ss} are evaluated in our FL environment, leveraging the abilities of the FedML library. To assess the effect of our sample selection methods across varying noise types, including open-set and closed-set noise, we train our MTAE models on MNIST and CIFAR10 datasets, both with and without sample selection, across different numbers of clients. Our primary goal is to improve global test accuracy, computed on the server, by leveraging the sample selection methods in the presence of noisy samples. We first conduct a series of training sessions without using sample selection and without adding noise to create a baseline for our experiments. The results of these experiments are presented in \cref{table:mnist_woval,table:cifar_woval}. The results show that as the total number of clients decreases, there is a corresponding increase in test accuracy. As the total number of clients decreases and the number of training samples is constant, the clients get larger training datasets during non-IID distribution. As the clients have larger local datasets with greater diversity, this results in better convergence of local models with increased accuracy and image quality.

\begin{table}[width=1\linewidth,cols=5,pos=h]
	\caption{The best performance of MTAE models trained on MNIST dataset without any noise and sample selection across varying numbers of clients.}
    \label{table:mnist_woval}
	\begin{tabular*}{\tblwidth}{@{} LLLLL@{} }
		\toprule
        Total Clients & Accuracy & PSNR & SSIM & Best Round \\
		\midrule
                1000  & 94.60 &	21.99 &	92.55 &	920 \\
                200   & 96.62 &	22.62 &	94.23 &	990 \\
                100   & 97.34 &	23.40 &	94.88 &	970 \\
                50    & 97.49 &	24.09 &	95.86 &	970 \\
		\bottomrule
	\end{tabular*}
\end{table}
\vspace{-1em}
\begin{table}[width=1\linewidth,cols=5,pos=h]
	\caption{The best performance of MTAE models trained on CIFAR10 dataset without any noise and sample selection across varying numbers of clients.}
    \label{table:cifar_woval}
	\begin{tabular*}{\tblwidth}{@{} LLLLL@{} }
		\toprule
        Total Clients & Accuracy & PSNR & SSIM & Best Round \\
		\midrule
                1000    & 57.95 & 18.54 & 43.92 & 960 \\
                200     & 66.81 & 20.15 & 57.60 & 999 \\
                100     & 69.75 & 20.87 & 62.08 & 970 \\
                50      & 71.05 & 21.55 & 66.26 & 900 \\
		\bottomrule
	\end{tabular*}
\end{table}

To establish a baseline for scenarios involving 40\% open-set and closed-set noise across varying numbers of clients, we performed a series of experiments on the CIFAR10 and MNIST datasets without using sample selection, similar to the experiments on clean datasets. The outcomes of these experiments are presented in \cref{table:mnist_noise,table:cifar_noise}. The closed-set experiments for both MNIST and CIFAR10 datasets show that the accuracy initially increases as the number of clients decreases, but then the accuracy starts to decrease after a certain point. This point is at 200 clients for MNIST and 100 clients for CIFAR10. This accuracy drop is due to increasing label noise on clients as the number of clients decreases, leading to instability in classification layers. 

In the MNIST experiments, closed-set noise has the most significant impact, reducing accuracy to 85.58\% for 50 clients. Additionally, using the letter images from the EMNIST dataset as open-set noise leads to a greater reduction in accuracy compared to ImageNet32, possibly due to the similarity of EMNIST and MNIST datasets. This shows that using random labels for similar images results in a reduction in the model's classification ability. The highest PSNR and SSIM are achieved for 50 clients under open-set noise, indicating successful reconstruction of test images. 

Like the MNIST experiments, closed-set noise causes the highest accuracy drop compared to open-set noise in the CIFAR10 experiments. The images in ImageNet32 also have a greater impact on the model than those from SVHN, regardless of the number of clients, possibly due to the similarity of CIFAR10 and ImagetNet32 datasets. The lowest accuracy observed, at 38.59\% for 50 clients, occurred under closed-set noise. The accuracy and image quality metrics, especially for ImageNet32, show improvement as the number of clients decreases. For the SVHN and closed-set noise, however, the accuracy begins to drop for fewer than 100 clients, possibly due to the increased presence of noisy samples on clients.

\begin{table}[width=1\linewidth,cols=6,pos=t]
	\caption{The best performance of MTAE models trained on MNIST dataset with 40\% noise and no sample selection across varying numbers of clients and noise types.}
    \label{table:mnist_noise}
	\begin{tabular*}{\tblwidth}{@{} LLLLLL@{} }
		\toprule
         Noise Type & Total Clients & Accuracy & PSNR & SSIM & Best Round           \\
		\midrule
        \multirow{4}{*}{Closed-Set} & 1000  & 90.14 &	21.38 &	91.56 &	640 \\
                                    & 200   & 92.28 &	21.42 &	91.86 &	470 \\
                                    & 100   & 88.36 &	22.40 &	92.84 &	460 \\
                                    & 50    & 85.58 &	21.96 &	92.69 &	260 \\
        \midrule
        \multirow{4}{*}{EMNIST}     & 1000  & 91.43 &   21.12 &	91.45 &	920 \\
                                    & 200   & 95.22 &   22.69 &	93.69 &	960 \\
                                    & 100   & 94.48 &   21.95 &	92.77 &	990 \\
                                    & 50    & 94.53 &   23.09 &	94.78 &	990 \\
        \midrule
        \multirow{4}{*}{ImageNet32}   & 1000  & 93.27 &	21.00 &	89.39 &	980 \\
                                    & 200   & 95.60 &	22.46 &	92.33 &	970 \\
                                    & 100   & 95.14 &	21.58 &	91.14 &	850 \\
                                    & 50    & 95.13 &	23.02 &	93.95 &	990 \\
		\bottomrule
	\end{tabular*}
\end{table}

\begin{table}[width=1\linewidth,cols=6,pos=t]
	\caption{The best performance of MTAE models trained on CIFAR10 dataset with 40\% noise and no sample selection across varying numbers of clients and noise types.}
    \label{table:cifar_noise}
	\begin{tabular*}{\tblwidth}{@{} LLLLLL@{} }
		\toprule
         Noise Type & Total Clients & Accuracy & PSNR & SSIM & Best Round           \\
		\midrule
        \multirow{4}{*}{Closed-Set} &    1000    & 39.74 & 18.04 & 40.77 & 820   \\
                                    &    200     & 40.32 & 19.78 & 55.04 & 980   \\
                                    &    100     & 41.36 & 20.29 & 59.71 & 890   \\
                                    &    50      & 38.59 & 20.26 & 59.47 & 550   \\

        \midrule
        \multirow{4}{*}{SVHN}       & 1000  & 50.99 & 18.16 &	41.21 &	970 \\
                                    & 200   & 57.46 & 19.54 &	53.30 &	970 \\
                                    & 100   & 60.86 & 20.13 &	58.23 &	930 \\
                                    & 50    & 60.58 & 20.49 &	60.91 &	900 \\

        \midrule
        \multirow{4}{*}{ImageNet32}   &    1000  &  47.47 &	18.52 &	43.67 &	990 \\
                                    &    200   &  52.68 &	19.84 &	55.04 &	890 \\
                                    &    100   &  55.25 &	20.43 &	60.39 &	890 \\
                                    &    50    &  57.53 &	20.90 &	63.06 &	900 \\
		\bottomrule
	\end{tabular*}
\end{table}

\subsection{Loss-based Selection} \label{sec_exp_loss_ss}

\newcommand{\fail}[1]{\underline{#1}}

\begin{table}[width=1\linewidth,cols=7,pos=t]
    \caption{The best performance of MTAE models trained on MNIST dataset with 40\% noise and sample selection across varying numbers of clients and noise types. The accuracy gains are shown in parentheses for easier evaluation, and underlining indicates that using sample selection leads to decreased accuracy.}
    \label{table:mnist_val}
	\begin{tabular*}{\tblwidth}{@{} LLLLLLL@{} }
		\toprule
        Noise Type               & Total Clients         & Method & Accuracy (Gain) & PSNR & SSIM & Best Round \\
        \midrule
        \multirow{12}{*}{Closed-Set} & \multirow{3}{*}{1000} & AT     & 91.75 ($+$1.61)       & 20.88 & 90.21 & 570        \\
                                     &                       & OCSVM  & 91.77 ($+$1.63)       & 21.29 & 91.36 & 780        \\
                                     &                       & IF     & 91.83 ($+$1.69)       & 21.27 & 91.37 & 770        \\ 
                                     \cmidrule{2-7} 
                                     & \multirow{3}{*}{200}  & AT     & 93.85 ($+$1.57)       & 21.13 & 91.21 & 510        \\
                                     &                       & OCSVM  &\fail{91.56} ($-$0.72)       & 21.60 & 92.32 & 700        \\
                                     &                       & IF     & 92.73 ($+$0.45)       & 21.62 & 92.36 & 770        \\ 
                                     \cmidrule{2-7} 
                                     & \multirow{3}{*}{100}  & AT     & 90.19 ($+$1.83)       & 22.38 & 92.77 & 550        \\
                                     &                       & OCSVM  & 88.71 ($+$0.35)       & 22.53 & 93.05 & 660        \\
                                     &                       & IF     & 89.82 ($+$1.46)       & 22.50 & 93.00 & 660        \\ 
                                     \cmidrule{2-7} 
                                     & \multirow{3}{*}{50}   & AT     & 85.58 ($+$0.00)       & 21.96 & 92.69 & 260        \\
                                     &                       & OCSVM  & 85.58 ($+$0.00)       & 21.96 & 92.69 & 260        \\
                                     &                       & IF     & 85.58 ($+$0.00)       & 21.96 & 92.69 & 260        \\ 
                                     \midrule
        \multirow{12}{*}{EMNIST}     & \multirow{3}{*}{1000} & AT     & 91.95 ($+$0.52)       & 20.91 & 90.95 & 880        \\
                                     &                       & OCSVM  & 92.06 ($+$0.63)       & 20.71 & 90.19 & 930        \\
                                     &                       & IF     &\fail{89.26} ($-$2.17)       & 20.62 & 89.97 & 880        \\ 
                                     \cmidrule{2-7} 
                                     & \multirow{3}{*}{200}  & AT     &\fail{94.99} ($-$0.23)       & 21.83 & 92.54 & 990        \\
                                     &                       & OCSVM  &\fail{94.03} ($-$1.19)       & 21.79 & 92.42 & 990        \\
                                     &                       & IF     &\fail{92.55} ($-$2.67)       & 21.82 & 92.50 & 990        \\ 
                                     \cmidrule{2-7} 
                                     & \multirow{3}{*}{100}  & AT     & 95.13 ($+$0.65)       & 22.49 & 93.38 & 960        \\
                                     &                       & OCSVM  & 95.06 ($+$0.58)       & 22.35 & 93.34 & 940        \\
                                     &                       & IF     &\fail{93.58} ($-$0.90)       & 22.52 & 93.59 & 999        \\ 
                                     \cmidrule{2-7} 
                                     & \multirow{3}{*}{50}   & AT     & 95.03 ($+$0.50)       & 22.77 & 94.26 & 860        \\
                                     &                       & OCSVM  & 95.65 ($+$1.12)       & 22.87 & 94.43 & 950        \\
                                     &                       & IF     &\fail{93.49} ($-$1.04)       & 22.50 & 93.87 & 610        \\ 
                                     \midrule
        \multirow{12}{*}{ImageNet32} & \multirow{3}{*}{1000} & AT     & 93.31 ($+$0.04)       & 20.77 & 88.65 & 980        \\
                                     &                       & OCSVM  & 93.47 ($+$0.20)       & 20.61 & 88.74 & 930        \\
                                     &                       & IF     &\fail{92.38} ($-$0.89)       & 20.83 & 89.46 & 950        \\ 
                                     \cmidrule{2-7} 
                                     & \multirow{3}{*}{200}  & AT     &\fail{95.32} ($-$0.28)       & 21.72 & 91.02 & 990        \\
                                     &                       & OCSVM  &\fail{95.35} ($-$0.25)       & 21.72 & 91.67 & 990        \\
                                     &                       & IF     &\fail{94.61} ($-$0.99)       & 21.91 & 92.20 & 990        \\ 
                                     \cmidrule{2-7} 
                                     & \multirow{3}{*}{100}  & AT     & 95.70 ($+$0.56)       & 22.18 & 91.36 & 960        \\
                                     &                       & OCSVM  & 95.97 ($+$0.83)       & 22.36 & 92.59 & 940        \\
                                     &                       & IF     & 95.27 ($+$0.13)       & 22.56 & 93.20 & 999        \\ 
                                     \cmidrule{2-7} 
                                     & \multirow{3}{*}{50}   & AT     & 95.54 ($+$0.41)       & 22.79 & 93.57 & 860        \\
                                     &                       & OCSVM  & 96.04 ($+$0.91)       & 22.96 & 93.90 & 950        \\
                                     &                       & IF     & 95.79 ($+$0.66)       & 23.34 & 94.59 & 970        \\
		\bottomrule
	\end{tabular*}
\end{table}

\begin{table}[width=1\linewidth,cols=7,pos=t]
	\caption{The best performance of MTAE models trained on CIFAR10 dataset with 40\% noise and sample selection across varying numbers of clients and noise types. The accuracy gains are shown in parentheses for easier evaluation, and underlining indicates that using sample selection leads to decreased accuracy.}
    \label{table:cifar_val}
	\begin{tabular*}{\tblwidth}{@{} LLLLLLL@{} }
		\toprule
        Noise Type                & Total Clients         & Method & Accuracy (Gain) & PSNR & SSIM & Best Round \\
        \midrule
        \multirow{12}{*}{Closed-Set} & \multirow{3}{*}{1000} & AT     & 40.94 ($+$1.20)          & 18.14 & 41.83 & 830        \\
                                     &                       & OCSVM  & 44.94 ($+$5.20)          & 18.33 & 43.17 & 999        \\
                                     &                       & IF     & 45.39 ($+$5.65)          & 18.29 & 43.12 & 999        \\ 
                                     \cmidrule{2-7}
                                     & \multirow{3}{*}{200}  & AT     &\fail{39.87} ($-$0.45)          & 19.56 & 53.56 & 880        \\ 
                                     &                       & OCSVM  & 46.76 ($+$6.44)          & 19.48 & 52.76 & 940        \\
                                     &                       & IF     & 46.17 ($+$5.85)          & 19.40 & 52.24 & 920        \\ 
                                     \cmidrule{2-7}
                                     & \multirow{3}{*}{100}  & AT     &\fail{40.48} ($-$0.88)          & 20.26 & 59.26 & 890        \\
                                     &                       & OCSVM  & 46.70 ($+$5.34)          & 20.03 & 57.61 & 930        \\
                                     &                       & IF     & 45.01 ($+$3.65)          & 19.92 & 57.06 & 990        \\ 
                                     \cmidrule{2-7}
                                     & \multirow{3}{*}{50}   & AT     & 38.60 ($+$0.01)          & 20.52 & 61.64 & 890        \\
                                     &                       & OCSVM  & 45.61 ($+$7.02)          & 20.43 & 60.85 & 890        \\
                                     &                       & IF     & 45.24 ($+$6.65)          & 20.42 & 60.47 & 890        \\ 
                                     \midrule
        \multirow{12}{*}{SVHN}       & \multirow{3}{*}{1000} & AT     &\fail{49.95} ($-$1.04)          & 18.16 & 41.17 & 990        \\
                                     &                       & OCSVM  & 51.40 ($+$0.41)          & 18.02 & 40.83 & 960        \\
                                     &                       & IF     &\fail{50.84} ($-$0.15)          & 17.88 & 40.08 & 940        \\ 
                                     \cmidrule{2-7}
                                     & \multirow{3}{*}{200}  & AT     &\fail{56.84} ($-$0.62)          & 19.47 & 52.70 & 999        \\
                                     &                       & OCSVM  &\fail{57.44} ($-$0.02)          & 19.14 & 50.58 & 999        \\ 
                                     &                       & IF     &\fail{56.82} ($-$0.64)          & 19.05 & 49.91 & 980        \\
                                     \cmidrule{2-7}
                                     & \multirow{3}{*}{100}  & AT     & 61.22 ($+$0.36)          & 20.03 & 57.66 & 930        \\
                                     &                       & OCSVM  & 60.86 ($+$0.00)          & 19.76 & 56.00 & 990        \\
                                     &                       & IF     &\fail{60.34} ($-$0.52)          & 19.70 & 55.56 & 990        \\ 
                                     \cmidrule{2-7}
                                     & \multirow{3}{*}{50}   & AT     & 61.14 ($+$0.56)          & 20.49 & 61.19 & 880        \\
                                     &                       & OCSVM  & 61.94 ($+$1.36)          & 20.21 & 59.01 & 890        \\
                                     &                       & IF     & 61.15 ($+$0.57)          & 20.33 & 60.54 & 990        \\ 
                                     \midrule
        \multirow{12}{*}{ImageNet32} & \multirow{3}{*}{1000} & AT     &\fail{47.00} ($-$0.47)          & 18.43 & 42.88 & 960        \\
                                     &                       & OCSVM  & 50.89 ($+$3.42)          & 18.43 & 43.27 & 999        \\
                                     &                       & IF     & 50.66 ($+$3.19)          & 18.38 & 43.08 & 999        \\ 
                                     \cmidrule{2-7}
                                     & \multirow{3}{*}{200}  & AT     &\fail{52.56} ($-$0.12)          & 19.90 & 55.69 & 999        \\
                                     &                       & OCSVM  & 55.99 ($+$3.31)          & 19.70 & 54.00 & 999        \\
                                     &                       & IF     & 56.29 ($+$3.61)          & 19.64 & 53.64 & 999        \\ 
                                     \cmidrule{2-7}
                                     & \multirow{3}{*}{100}  & AT     &\fail{53.37} ($-$1.88)          & 20.24 & 59.02 & 890        \\
                                     &                       & OCSVM  & 57.79 ($+$2.54)          & 20.14 & 58.04 & 930        \\
                                     &                       & IF     & 58.72 ($+$3.47)          & 20.20 & 58.58 & 970        \\ 
                                     \cmidrule{2-7}
                                     & \multirow{3}{*}{50}   & AT     & 58.09 ($+$0.56)          & 21.08 & 63.49 & 940        \\
                                     &                       & OCSVM  & 59.04 ($+$1.51)          & 20.62 & 61.75 & 890        \\
                                     &                       & IF     & 59.02 ($+$1.49)          & 20.63 & 62.28 & 900        \\
		\bottomrule
	\end{tabular*}
\end{table}

In this work, we present a novel approach to loss-based sample selection tailored specifically for federated learning. We use OCSVM, IF, and AT methods within the FL context, explained in detail in \cref{sec_at,sec_uod}. During training, we evaluate these methods by introducing various types of noise to the training dataset. We use a high noise rate of 40\% to better observe the benefits of sample selection. The methods are updated on the server every 5 rounds using the weighted losses collected from each client. The update period should be chosen carefully as it significantly impacts the total training time, given the computational cost of training OCSVM and IF models. If the update period is chosen too long, the sample selection methods may fail to adapt to the loss distribution of the current FL model, potentially leading to the selection of noisy samples. 

We optimize the loss weights $\lambda_{\text{rec}}$ and $\lambda_{\text{cls}}$ through extensive experiments on MNIST and CIFAR10 datasets. The loss weights $\lambda_{\text{rec}}=1$ and $\lambda_{\text{cls}}=0.05$ are empirically chosen to balance reconstruction and classification tasks. We observe that a higher weight on reconstruction ensures better image quality and helps identify noisy samples, while a lower classification weight prevents overfitting and maintains PSNR and SSIM scores. Since we use a noise rate of 40\%, we set the contamination parameter to 0.4 for both OCSVM and IF. We use the square root of the data size as the number of estimators for the IF algorithm to balance computational efficiency and model accuracy, ensuring that the model captures sufficient data variation without excessive overhead. For OCSVM, we use the radial basis function (RBF) kernel as it is well-suited for handling non-linear data, enabling better separation of outliers from normal samples in high-dimensional feature spaces. 

We optimize the parameters of the AT through extensive experiments on the MNIST and CIFAR10 datasets, using FedBalancer’s parameters as a reference \cite{shin2022fedbalancer}. We select a loss step size of $lss=0.1$ for smooth threshold adjustments and an update period of $t_w=5$ to adapt to the new loss distribution, aligning with the update period of OCSVM and IF models. The probability $p=0.75$ is set to retain high-loss samples, which could either be valuable or noisy. Since the AT method relies on a single loss function, we use the final loss, which is the weighted combination of CE and MSE losses.

We activate our sample selection methods after the 400th round, as this is when the rate of accuracy improvement begins to slow, signaling that the model is stabilizing. Choosing the right starting point for sample selection is a crucial hyperparameter as it greatly influences the method's effectiveness. In the early rounds, underfitting results in uniformly high losses, while in the later rounds, overfitting causes noisy samples to be memorized, leading to lower losses. Both underfitting and overfitting make the losses less distinctive, complicating the differentiation between valuable and noisy samples.

The results of experiments conducted on the MNIST and CIFAR10 datasets are shown in \cref{table:mnist_val,table:cifar_val}. Underlined results show that using the corresponding sample selection method worsens model accuracy compared to \cref{table:mnist_noise,table:cifar_noise}, while the others show accuracy improvements. In the MNIST experiments with closed-set noise, AT generally achieves the highest accuracy in shorter rounds, with the most improvement of 1.83\% for 100 clients. IF improves accuracy for 1000, 200, and 100 clients, with the most improvement of 1.69\% for 1000 clients. OCSVM also improves accuracy for 1000 and 100 clients, with the highest improvement of 1.63\% for 1000 clients. However, OCSVM reduces accuracy by 0.72\% for 200 clients, likely due to the RBF kernel's tendency to overfit, leading to the misclassification of outliers.

In the EMNIST experiments, OCSVM surpasses IF and AT in accuracy, with the highest improvement of 1.12\% for 50 clients. IF fails across all numbers of clients, likely due to the domain similarity of MNIST and EMNIST, resulting in overlapping between normal and noisy samples. However, in the ImageNet32 open-set experiments, IF performs better and improves accuracy for 50 and 100 clients, with the highest gain of 0.66\% for 50 clients. OCSVM achieves the highest overall gain in this setting, with a 0.91\% accuracy gain for 50 clients. AT improves accuracy for all numbers of clients except 200, where all methods show a reduction, likely due to the delayed initiation of sample selection, leading to overfitting on noisy samples and the misclassification of outliers.

In the CIFAR10 experiments with closed-set noise, OCSVM and IF enhance accuracy across all numbers of clients, while AT improves accuracy by 1.2\% and 0.01\% for 1000 and 50 clients, respectively. However, AT slightly decreases accuracy for 100 and 200 clients, with decent image quality. The highest improvement, which is 7.02\% for 50 clients, is achieved by OCSVM. IF also shows significant improvement, with the greatest increase of 6.65\% for 50 clients. The results show that our methods filter noisy samples successfully, allowing stable training and high accuracy. 

In the CIFAR10 experiments with SVHN open-set noise, OCSVM improves accuracy by 0.41\% for 1000 clients and 1.36\% for 50 clients but shows no improvement for 100 and 200 clients. IF is only effective for 50 clients, increasing accuracy by 0.57\%. AT reduces accuracy for 1000 and 200 clients but provides slight improvements of 0.36\% and 0.56\% for 100 and 50 clients, respectively. These results are likely due to the model's memorization of SVHN samples, which is a simpler dataset compared to ImageNet32, leading to low loss values that hinder our methods from effectively distinguishing noisy samples. In the ImageNet32 experiments, OCSVM and IF show significant accuracy improvement for all numbers of clients, while AT increases accuracy by 0.56\% for only 50 clients. The greatest improvement is achieved by IF, which is 3.61\% for 200 clients. Overall, AT struggles to improve accuracy as the client number increases, likely due to the non-IID nature of data across clients, leading to variations in loss values and complicating the application of a global threshold to distinguish noisy samples. OCSVM proves its effectiveness in almost all scenarios, while IF performs well overall, except when faced with open-set noise from the SVHN dataset.

\begin{table*}[width=1.0\textwidth,cols=12,pos=t]
    \centering
    \caption{The best performance of MTAE models trained on MNIST and CIFAR10 datasets with 40\% noise for 1000 clients, evaluated using macro-averaged classification metrics. The performance gains are shown in parentheses for easier evaluation, and underlining indicates that using sample selection leads to reduced performance.}
    \label{table:mnist_cifar_val_cls}
	\begin{tabular*}{\tblwidth}{@{} LLLLLLLLLLLL@{} }
        \toprule
        \multicolumn{6}{c}{MNIST} & \multicolumn{6}{c}{CIFAR10} \\
        \cmidrule{1-6} \cmidrule{7-12}
        Noise Type & Method & Precision (Gain) & Recall (Gain) & F1 Score (Gain) & Best Round & Noise Type & Method & Precision (Gain) & Recall (Gain) & F1 Score (Gain) & Best Round \\
        \midrule
        \multirow{4}{*}{Closed-Set} & --      & 90.23 & 89.98 & 89.98 & 640 & \multirow{4}{*}{Closed-Set} & --        & 40.56 & 39.74 & 39.15 & 820 \\
                                    & AT      & 91.86 ($+$1.63) & 91.62 ($+$1.64) & 91.67 ($+$1.69) & 570 & & AT      & 40.65 ($+$0.09) & 40.94 ($+$1.20) & 40.41 ($+$1.26) & 830 \\
                                    & OCSVM   & 91.88 ($+$1.65) & 91.70 ($+$1.72) & 91.66 ($+$1.68) & 780 & & OCSVM   & 46.11 ($+$5.55) & 44.94 ($+$5.20) & 44.79 ($+$5.64) & 999 \\
                                    & IF      & 91.85 ($+$1.62) & 91.77 ($+$1.79) & 91.74 ($+$1.76) & 770 & & IF      & 46.14 ($+$5.58) & 45.39 ($+$5.65) & 45.00 ($+$5.85) & 999 \\
        \cmidrule{1-6} \cmidrule{7-12}
        \multirow{4}{*}{EMNIST}     & --      & 91.73 & 91.28 & 91.36 & 920 & \multirow{4}{*}{SVHN}       & --        & 50.69 & 50.99 & 50.51 & 970 \\
                                    & AT      & 92.14 ($+$0.41) & 91.93 ($+$0.65) & 91.92 ($+$0.56) & 880 & & AT      & \fail{49.97} ($-$0.72) & \fail{49.95} ($-$1.04) & \fail{49.41} ($-$1.10) & 990 \\
                                    & OCSVM   & 92.06 ($+$0.33) & 91.99 ($+$0.71) & 92.01 ($+$0.65) & 930 & & OCSVM   & 50.70 ($+$0.01) & 51.40 ($+$0.41) & 50.61 ($+$0.10) & 960 \\
                                    & IF      & \fail{89.36} ($-$2.37) & \fail{89.20} ($-$2.08) & \fail{89.20} ($-$2.16) & 880 & & IF      & 50.90 ($+$0.21) & \fail{50.84} ($-$0.15) & \fail{50.30} ($-$0.21) & 940 \\
        \cmidrule{1-6} \cmidrule{7-12}
        \multirow{4}{*}{ImageNet32} & --      & 93.43 & 93.19 & 93.23 & 980 & \multirow{4}{*}{ImageNet32} & --        & 48.23 & 47.47 & 47.02 & 990 \\
                                    & AT      & 93.50 ($+$0.07) & 93.22 ($+$0.03) & 93.27 ($+$0.04) & 980 & & AT      & \fail{46.39} ($-$1.84) & \fail{47.00} ($-$0.47) & \fail{45.85} ($-$1.17) & 960 \\
                                    & OCSVM   & 93.44 ($+$0.01) & 93.45 ($+$0.26) & 93.43 ($+$0.20) & 930 & & OCSVM   & 51.15 ($+$2.92) & 50.89 ($+$3.42) & 50.15 ($+$3.13) & 999 \\
                                    & IF      & \fail{92.41} ($-$1.02) & \fail{92.30} ($-$0.89) & \fail{92.32} ($-$0.91) & 950 & & IF      & 50.76 ($+$2.53) & 50.66 ($+$3.19) & 49.82 ($+$2.80) & 999 \\
        \bottomrule
    \end{tabular*}
\end{table*}

In addition to accuracy, we further evaluate our sample selection methods using macro-averaged precision, recall, and F1 score metrics, which provide a more balanced view of classification performance across 1000 clients, as shown in \ref{table:mnist_cifar_val_cls}. Macro-averaging computes the metric for each class and averages them, giving equal weight to all classes, regardless of size. The results indicate that OCSVM consistently outperforms the other methods, especially on CIFAR10, with significant improvements in all three metrics. For example, OCSVM shows a gain of 5.55\% in precision and 5.64\% in F1 score compared to the baseline under closed-set noise. AT performs well on MNIST, but its effectiveness diminishes on CIFAR10 with SVHN noise, occasionally underperforming relative to the baseline. In contrast, IF delivers mixed results, excelling on CIFAR10 but showing notable declines on MNIST with EMNIST noise, where all three metrics drop by over 2\%. Overall, OCSVM emerges as the most reliable method across various noise types and datasets, and these results closely align with those from the accuracy metric, demonstrating consistent trends in performance.

\subsection{Feature-based Selection} \label{sec_exp_ftr_ss}

We propose a feature-based sample selection approach using the OCSVM and IF algorithms described in \cref{sec_uod}, with the same configuration as the loss-based sample selection experiments. During training, we introduce 40\% open-set noise to the training set while excluding closed-set noise, as the latter does not present detectable outliers in the feature space. As in loss-based sample selection, OCSVM and IF models are trained every 5 rounds and used to identify outliers in the subsequent 5 training rounds. Additionally, we incorporate our federated multi-class SVDD loss to enhance the distinctiveness of embeddings for the feature-based sample selection, described in \cref{sec_fedsvdd}. The proposed SVDD loss is added to the final client loss with a regularization weight of $\lambda_{\text{reg}}=10^{-5}$, which is empirically chosen to maintain training stability while effectively incorporating the SVDD objective. Higher values are found to distort the embedding space, reducing the method’s overall effectiveness. 

We activate our feature-based sample selection process after the 600th round of training, while the federated SVDD loss is applied earlier, starting at the 500th round. This staggered timing allows the model first to enhance the distinctiveness of embeddings through the SVDD loss, ensuring better separation of normal and outlier samples. Delaying the sample selection by 200 rounds, compared to the loss-based approach, provides the model more time to learn meaningful feature representations, minimizing the risk of misclassifying valuable samples as outliers. This approach ensures sufficiently refined embedding space, leading to more precise and effective sample selection.

The results of the experiments performed on the CIFAR10 and MNIST datasets are shown in \cref{table:exp_ss_ftr_all}. The experimental results show that the feature-based sample selection with or without the SVDD loss reduces the model accuracy for all scenarios when compared to the results of baseline experiments in \cref{table:mnist_noise,table:cifar_noise}. These findings indicate that identifying noisy samples in a feature space of 512 or 1024 dimensions is not as effective as expected. This could be possibly due to the high noise rate and the memorization of noisy samples, making them indistinguishable in the feature space. In the CIFAR10 experiments, however, it is notable that our federated SVDD loss improves the effectiveness of feature-based sample selection in terms of accuracy and image quality for large numbers of clients, such as 200 and 1000, compared to the experiments without it. We believe that these findings are important for guiding future studies on feature-based sample selection.

\begin{table*}[width=1.0\textwidth,cols=12,pos=t]
	\caption{The best performance of MTAE models trained on CIFAR10 and MNIST datasets with 40\% noise and feature-based sample selection across varying numbers of clients. The underline indicates that feature-based sample selection decreases accuracy when the federated SVDD loss is utilized after the 500th round.}
    \label{table:exp_ss_ftr_all}
	\begin{tabular*}{\tblwidth}{@{} LLLLLLLLLLLL@{} }
	\toprule        
        \multirow{2}{*}{Train Dataset} & \multirow{2}{*}{Noise Type} & \multirow{2}{*}{Total Clients} & \multirow{2}{*}{Method} & \multicolumn{4}{c}{Without SVDD}  & \multicolumn{4}{c}{With SVDD}           \\ 
        \cline{5-8} \cline{9-12} \rule{0pt}{2.5ex} 
                                        & & & & Accuracy & PSNR  & SSIM  & Best Round & Accuracy (Gain) & PSNR  & SSIM  & Best Round \\
        \midrule   
        \multirow{16}{*}{CIFAR10}       & \multirow{8}{*}{SVHN}           & \multirow{2}{*}{1000} & OCSVM    & 47.24 & 17.49 & 36.74 & 930 & 47.64 ($+$0.40) & 17.67 & 38.16 & 980 \\
                                        &                                 &                       & IF       & 47.95 & 17.60 & 38.10 & 960 & 48.36 ($+$0.41) & 17.68 & 38.23 & 960 \\
                                        \cmidrule{3-12}
                                        &                                 & \multirow{2}{*}{200}  & OCSVM    & 53.60 & 18.45 & 47.90 & 800 & 54.19 ($+$0.59) & 18.80 & 49.02 & 999 \\
                                        &                                 &                       & IF       & 53.07 & 18.25 & 47.16 & 690 & 53.51 ($+$0.44) & 18.85 & 49.00 & 940 \\
                                        \cmidrule{3-12}
                                        &                                 & \multirow{2}{*}{100}  & OCSVM    & 58.02 & 19.49 & 54.68 & 930 & \fail{57.43} ($-$0.59) & 19.52 & 53.93 & 600 \\
                                        &                                 &                       & IF       & 58.07 & 19.46 & 54.17 & 720 & \fail{57.43} ($-$0.64) & 19.52 & 53.93 & 600 \\
                                        \cmidrule{3-12}
                                        &                                 & \multirow{2}{*}{50}   & OCSVM    & 60.24 & 20.11 & 58.43 & 940 & \fail{58.30} ($-$1.94) & 19.90 & 57.34 & 550 \\
                                        &                                 &                       & IF       & 58.72 & 19.96 & 58.24 & 720 & \fail{58.30} ($-$0.42) & 19.90 & 57.34 & 550 \\
        \cmidrule{2-12}
                                        & \multirow{8}{*}{ImageNet32}     & \multirow{2}{*}{1000} & OCSVM    & 46.81 &	17.87 &	39.05 &	820 & 46.98 ($+$0.17) & 18.11 & 40.88 & 930 \\
                                        &                                 &                       & IF       & 46.82 &	18.07 &	40.66 &	960 & 47.30 ($+$0.48) & 18.21 & 41.63 & 990 \\
                                        \cmidrule{3-12}
                                        &                                 & \multirow{2}{*}{200}  & OCSVM    & 50.43 &	18.98 &	50.68 &	690 & 51.42 ($+$0.99) & 19.37 & 52.70 & 970 \\
                                        &                                 &                       & IF       & 51.07 &	19.17 &	52.46 &	999 & \fail{49.45} ($-$1.62) & 19.38 & 52.70 & 999 \\
                                        \cmidrule{3-12}
                                        &                                 & \multirow{2}{*}{100}  & OCSVM    & 54.48 &	20.08 &	58.15 &	930 & \fail{53.06} ($-$1.42) & 19.99 & 57.70 & 930 \\
                                        &                                 &                       & IF       & 52.44 &	19.99 &	56.75 &	600 & 53.24 ($+$0.80) & 19.88 & 56.90 & 970 \\
                                        \cmidrule{3-12}
                                        &                                 & \multirow{2}{*}{50}   & OCSVM    & 56.04 &	20.26 &	59.49 &	900 & \fail{55.41} ($-$0.63) & 20.66 & 61.78 & 940 \\
                                        &                                 &                       & IF       & 55.28 &	20.68 &	61.52 &	940 & \fail{54.20} ($-$1.08) & 20.47 & 60.87 & 940 \\
        \midrule
        \multirow{16}{*}{MNIST}         & \multirow{8}{*}{EMNIST}         & \multirow{2}{*}{1000} & OCSVM    & 89.40 &	20.09 &	89.02 &	920 & \fail{88.90} ($-$0.50) & 20.00 & 87.71 & 540 \\
                                        &                                 &                       & IF       & 89.11 &	20.16 &	88.41 &	540 & \fail{88.90} ($-$0.21) & 20.00 & 87.71 & 540 \\
                                        \cmidrule{3-12}
                                        &                                 & \multirow{2}{*}{200}  & OCSVM    & 92.90 &	21.20 &	91.21 &	600 & \fail{92.59} ($-$0.31) & 21.08 & 90.99 & 600 \\
                                        &                                 &                       & IF       & 92.90 &	21.20 &	91.21 &	600 & \fail{92.59} ($-$0.31) & 21.08 & 90.99 & 600 \\
                                        \cmidrule{3-12}
                                        &                                 & \multirow{2}{*}{100}  & OCSVM    & 92.99 &	21.49 &	91.77 &	670 & \fail{92.33} ($-$0.66) & 21.23 & 91.02 & 670 \\
                                        &                                 &                       & IF       & 92.32 &	21.21 &	91.22 &	960 & \fail{91.60} ($-$0.72) & 21.19 & 91.06 & 420 \\
                                        \cmidrule{3-12}
                                        &                                 & \multirow{2}{*}{50}   & OCSVM    & 92.86 &	21.52 &	92.02 &	330 & 92.86 ($+$0.00) & 21.52 & 92.02 & 330 \\
                                        &                                 &                       & IF       & 94.32 &	22.20 &	93.41 &	610 & \fail{93.05} ($-$1.27) & 21.80 & 92.59 & 610 \\
        \cmidrule{2-12}        
                                        & \multirow{8}{*}{ImageNet32}     & \multirow{2}{*}{1000} & OCSVM    & 93.04 &	20.26 &	87.43 &	930 & \fail{92.02} ($-$1.02) & 19.63 & 82.82 & 920 \\
                                        &                                 &                       & IF       & 91.91 &	19.77 &	83.89 &	930 & \fail{91.22} ($-$0.69) & 19.24 & 80.67 & 920 \\
                                        \cmidrule{3-12}
                                        &                                 & \multirow{2}{*}{200}  & OCSVM    & 94.14 &	21.05 &	89.88 &	900 & 94.29 ($+$0.15) & 20.85 & 88.64 & 990 \\
                                        &                                 &                       & IF       & 94.06 &	20.95 &	89.67 &	600 & \fail{93.30} ($-$0.76) & 20.57 & 87.36 & 600 \\
                                        \cmidrule{3-12}
                                        &                                 & \multirow{2}{*}{100}  & OCSVM    & 94.56 &	21.53 &	89.87 &	960 & \fail{93.83} ($-$0.73) & 21.30 & 87.83 & 960 \\
                                        &                                 &                       & IF       & 93.41 &	21.06 &	87.11 &	960 & \fail{92.55} ($-$0.86) & 20.95 & 89.04 & 420 \\
                                        \cmidrule{3-12}
                                        &                                 & \multirow{2}{*}{50}   & OCSVM    & 95.27 &	22.05 &	91.82 &	950 & \fail{93.87} ($-$1.40) & 21.93 & 92.00 & 460 \\
                                        &                                 &                       & IF       & 95.59 &	22.13 &	92.12 &	610 & \fail{93.87} ($-$1.72) & 21.93 & 92.00 & 460 \\
		\bottomrule
	\end{tabular*}
\end{table*}

\begin{table*}[width=1.0\textwidth,cols=11,pos=t]
	\caption{The best performance of MTAE models trained on CIFAR10 and MNIST datasets with 40\% noise and feature-based sample selection for 1000 clients, with additional classification metrics. The underline shows that using our multi-class federated SVDD loss after the 500th round reduces performance compared to not using it.}
    \label{table:ss_ftr_cls}
	\begin{tabular*}{\tblwidth}{@{} LLLLLLLLLLL@{} }
	\toprule        
        \multirow{2}{*}{Train Dataset} & \multirow{2}{*}{Noise Type} & \multirow{2}{*}{Method} & \multicolumn{4}{c}{Without SVDD}  & \multicolumn{4}{c}{With SVDD}           \\ 
        \cline{4-7} \cline{8-11} \rule{0pt}{2.5ex} 
                                        & & & Precision & Recall  & F1 Score & Best Round & Precision (Gain) & Recall (Gain) & F1 Score (Gain) & Best Round \\
        \midrule   
        \multirow{4}{*}{CIFAR10}        & \multirow{2}{*}{SVHN}           & OCSVM    & 46.74 & 47.24 & 46.72 & 930 & 47.81 ($+$1.07) & 47.64 ($+$0.40) & 47.30 ($+$0.58) & 980 \\
                                        &                                 & IF       & 47.20 & 47.95 & 47.19 & 960 & 47.60 ($+$0.40) & 48.36 ($+$0.41) & 47.37 ($+$0.18) & 960 \\
        \cmidrule{2-11}
                                        & \multirow{2}{*}{ImageNet32}     & OCSVM    & 46.40 & 46.81 & 46.14 & 820 & 47.12 ($+$0.72) & 46.98 ($+$0.17) & 46.69 ($+$0.55) & 930 \\
                                        &                                 & IF       & 46.21 & 46.82 & 45.76 & 960 & 48.02 ($+$1.81) & 47.30 ($+$0.48) & 46.62 ($+$0.86) & 990 \\
        \midrule
        \multirow{4}{*}{MNIST}          & \multirow{2}{*}{EMNIST}         & OCSVM    & 89.77 & 89.27 & 89.37 & 920 & \fail{89.09} ($-$0.68) & \fail{88.74} ($-$0.53) & \fail{88.76} ($-$0.61) & 540 \\
                                        &                                 & IF       & 89.21 & 88.97 & 88.98 & 540 & \fail{89.09} ($-$0.12) & \fail{88.74} ($-$0.23) & \fail{88.76} ($-$0.22) & 540 \\
        \cmidrule{2-11}
                                        & \multirow{2}{*}{ImageNet32}     & OCSVM    & 93.08 & 93.01 & 93.00 & 930 & \fail{92.15} ($-$0.93) & \fail{91.98} ($-$1.03) & \fail{92.00} ($-$1.00) & 920 \\
                                        &                                 & IF       & 91.97 & 91.84 & 91.85 & 930 & \fail{91.34} ($-$0.63) & \fail{91.13} ($-$0.71) & \fail{91.15} ($-$0.70) & 920 \\
		\bottomrule
	\end{tabular*}
\end{table*}

In the CIFAR10 experiments with SVHN open-set noise, our federated SVDD loss is proven to be effective for 1000 and 200 clients, with OCSVM achieving the highest improvement of 0.59\% for 200 clients. Similarly, in the ImageNet32 experiments, we observe that it is also effective for 1000, 200, and 100 clients, with the highest accuracy improvement of 0.99\%, acquired by OCSVM for 200 clients. However, no accuracy improvements are observed for 100 and 50 clients in the SVHN experiments, nor for 50 clients in the ImageNet32 experiments. Additionally, in the ImageNet32 experiments, our SVDD loss restricts the effectiveness of OCSVM for 100 clients and IF for 200 clients. This is possibly due to overfitting and pushing the embedding of noisy samples to class centroids, reducing classification accuracy and making it challenging to identify noisy samples in the feature space. Likely due to the same reason, which will be thoroughly examined in future work, our federated SVDD loss shows no significant accuracy improvement across all scenarios in the MNIST experiments. We also observe a reduction in the image quality, measured by PSNR and SSIM. This indicates the need for hyperparameter optimization of our method, including the starting round and the loss weight $\lambda_{\text{reg}}$, which is also a subject for future investigation.

The results presented in \cref{table:ss_ftr_cls} provide a more detailed analysis of the classification performance for feature-based sample selection with 1000 clients in terms of precision, recall, and F1 score, which are calculated in a macro-averaged mode. Similar to accuracy-based observations in \cref{table:exp_ss_ftr_all}, when the federated SVDD loss is applied, OCSVM shows an improvement in F1 score of 0.58\% for the CIFAR10 dataset with SVHN noise and a 0.55\% increase for the ImageNet32 dataset with the same noise. Similarly, IF achieves a notable gain of 0.86\% in the F1 score for ImageNet32. This suggests that the federated SVDD loss helps enhance the feature-based sample selection in complex datasets with high variability like CIFAR10. However, for the MNIST dataset with EMNIST noise, the federated SVDD loss results in a decrease in the F1 score, indicating a negative impact on classification performance. In these cases, the overfitting issue mentioned earlier might have contributed to the drop in performance, pushing noisy sample embeddings closer to class centroids and making outlier detection less effective. This indicates the importance of carefully timing the application of the SVDD loss, considering the dataset complexity and the noise characteristics.

\section{Limitations}

Despite their promising results across various client counts and noise types, our sample selection methods have certain limitations, primarily due to the computational complexity of outlier detection algorithms like OCSVM and IF. For the server, OCSVM has a training complexity ranging from $\mathcal{O}(n^2 \cdot d)$ to $\mathcal{O}(n^3 \cdot d)$ \cite{libsvm}, depending on convergence speed and the number of support vectors, while IF has $\mathcal{O}(t \cdot \psi \log \psi)$ \cite{liu2008isolation}, where $n$ is the number of samples, $d$ is the dimensionality of the data, $\psi$ is the sub-sampling size, and $t$ is the number of trees. For a client $i$ with $n_i$ samples, the prediction complexity is $\mathcal{O}(n_i \cdot SV \cdot d)$ for OCSVM, where $SV$ is the number of support vectors, and $\mathcal{O}(n_i \cdot t \log \psi)$ for IF. For our SVDD loss, the time complexity for a client $i$ is $\mathcal{O}(n_i \cdot d)$ due to the distance calculation for each sample. On the server, the centroids are computed once on a test set of size $n_t$ with a complexity of $\mathcal{O}(k \cdot n_t \cdot d)$, where $k$ is the number of classes. For each round, the radii are updated with a complexity of $\mathcal{O}(k \cdot n_d \log n_d)$, as it involves sorting the collected distances from clients, where $n_d$ is the total number of distances. In contrast, the AT method has negligible communication and computation costs for clients, requiring only minimal metadata and simple threshold-based filtering.

Another limitation is that the contamination parameters of IF and OCSVM must be properly set according to the noise level in the dataset; if set too high, high-quality samples risk being misclassified as outliers, which can reduce the accuracy of the model. To address this, we could use adaptive methods to adjust the contamination parameter by monitoring performance metrics or loss statistics. Our sample selection methods require careful timing during training, as applying it too late may result in the memorization of noisy samples, reducing their effectiveness. To mitigate this, we could use adaptive strategies to determine an optimal starting round for our methods, guided by indicators of model convergence or loss stability. Additionally, our AT method could be improved by using multiple loss functions and thresholds, rather than relying on a single one, to determine outlier samples more effectively, especially for complex datasets such as CIFAR10. Furthermore, we could improve our federated SVDD algorithm by incorporating class weights on clients to address the unbalanced nature of non-IID distributions, leading to better separation of outliers from normal samples in feature-based sample selection.

\section{Conclusion}

In this paper, we introduced novel sample selection methods to improve the effectiveness of FL models during training without requiring any preprocessing steps or pre-trained models. Our approach includes unsupervised outlier detection and adaptive threshold methods, managed by a central server, to select high-quality data samples during training on clients. Our approach effectively estimates sample contributions through loss and feature analysis by leveraging a multi-task autoencoder architecture for image classification and reconstruction. 

We developed a multi-process simulation environment using the FedML library to evaluate our methods. We proposed novel methods for loss-based sample selection by using OCSVM, IF, and AT methods explicitly tailored for FL. These methods are designed to periodically update themselves using losses or features collected from clients during training on non-IID datasets. The experimental results showed significant improvements in global test accuracy, precision, recall, and F1 score. Notably, we achieved accuracy gains of up to 7.02\% on CIFAR10 with OCSVM and 1.83\% on MNIST with AT across various noise settings and numbers of clients. Although IF achieves the highest F1 gains, such as 5.85\% on CIFAR10 and 1.76\% on MNIST for 1000 clients, OCSVM generally delivers more consistent performance across various noise settings. Additionally, we proposed a multi-class deep SVDD algorithm designed explicitly for FL to enhance feature-based sample selection performance through OCSVM and IF. However, its effectiveness was only confirmed for CIFAR10, with modest accuracy gains of up to 0.99\% using OCSVM and 0.80\% with IF. 

In future work, we will address the computational overhead of the proposed methods by exploring lightweight and scalable solutions that maintain high performance while reducing the FL resource demands in large-scale real-world scenarios. We aim to explore optimal activation strategies for sample selection and adaptive techniques for dynamically adjusting hyperparameters, such as the contamination rates in OCSVM and IF, to enhance robustness against varying noise levels without manual tuning. We also plan to expand our methods to tackle more complex datasets and investigate their applicability in real-world FL scenarios with heterogeneous devices and non-IID data distributions.

\pdfbookmark[section]{CRediT authorship contribution statement}{}
\printcredits

\pdfbookmark[section]{Declaration of competing interest}{}
\section*{Declaration of competing interest}
The authors declare that they have no known competing financial
interests or personal relationships that could have appeared
to influence the work reported in this paper.

\pdfbookmark[section]{Acknowledgment}{} 
\section*{Acknowledgment}
We sincerely thank Prof. Fatih Erdoğan Sevilgen and Assoc. Prof. Mehmet Göktürk for their valuable insights and constructive feedback. Their guidance has significantly contributed to the quality of this work. We also thank the editors and anonymous reviewers for their thoughtful suggestions, which have helped enhance this research paper.

\pdfbookmark[section]{Declaration of Generative AI and AI-assisted technologies in the writing process}{}
\section*{Declaration of Generative AI and AI-assisted technologies in the writing process}
During the preparation of this work, the author(s) used OpenAI's ChatGPT-3.5/4o to improve readability and language. After using this tool/service, the author(s) reviewed and edited the content as needed and take(s) full responsibility for the content of the publication.

\pdfbookmark[section]{References}{} 
\hyphenpenalty=10000 
\bibliographystyle{elsarticle-num} 				
\bibliography{fl-refs}

\vskip6pt

\end{document}